\documentclass[10pt,twocolumn,letterpaper]{article}

\usepackage[pagenumbers]{cvpr} 

\usepackage[dvipsnames]{xcolor}
\newcommand{\red}[1]{{\color{red}#1}}

\definecolor{cvprblue}{rgb}{0.21,0.49,0.74}
\usepackage[pagebackref,breaklinks,colorlinks,citecolor=cvprblue]{hyperref}

\usepackage{times}
\usepackage{epsfig}
\usepackage{graphicx}
\usepackage{amsmath}
\usepackage{amssymb}
\usepackage{multirow} 
\usepackage{makecell}
\usepackage{caption}
\usepackage{subcaption}
\usepackage{stackengine}
\usepackage{booktabs}
\usepackage[dvipsnames]{xcolor}
\usepackage{enumitem}
\usepackage{algorithm}
\usepackage{algorithmic}
\usepackage{afterpage}

\newcommand{\project}{MAP-ADAPT}
\newcommand{\green}[1]{\textcolor{ForestGreen}{#1}}
\newcommand{\blue}[1]{\textcolor{blue}{#1}}
\newcommand{\grey}[1]{\textcolor{gray}{#1}}

\definecolor{eccvblue}{rgb}{0.21,0.49,0.74}
\definecolor{eccvorange}{rgb}{1.0,0.49,0.0}
\definecolor{eccvgreen}{rgb}{0.0,0.5,0.0}

\def\paperID{*****} 
\def\confName{CVPR}
\def\confYear{2024}

\title{MAP-ADAPT: Real-Time Quality-Adaptive Semantic 3D Maps}

\author{Jianhao Zheng$^{1}$ \quad
Daniel Barath$^{2}$ \quad
Marc Pollefeys$^{2,3}$ \quad
Iro Armeni$^{1}$\\
$^{1}$Stanford Univsersity \quad $^{2}$ETH Zurich \quad $^{3}$Microsoft \\
}

\begin{document}
\maketitle
\begin{abstract}
Creating 3D semantic reconstructions of environments is fundamental to many applications, especially when related to autonomous agent operation (\eg, goal-oriented navigation or object interaction and manipulation). Commonly, 3D semantic reconstruction systems capture the entire scene in the same level of detail. However, certain tasks (\eg, object interaction) require a fine-grained and high-resolution map, particularly if the objects to interact are of small size or intricate geometry. In recent practice, this leads to the entire map being in the same high-quality resolution, which results in increased computational and storage costs. To address this challenge, we propose \textit{\project{}}, a real-time method for quality-adaptive semantic 3D reconstruction using RGBD frames. \project{} is the first adaptive semantic 3D mapping algorithm that, unlike prior work, generates directly a \textit{single} map with regions of different quality based on both the semantic information and the geometric complexity of the scene. Leveraging a semantic SLAM pipeline for pose and semantic estimation, we achieve comparable or superior results to state-of-the-art methods on synthetic and real-world data, while significantly reducing storage and computation requirements.
\end{abstract}    
\section{Introduction}
\label{sec:intro}

\begin{figure}[tb]
    \centering
    \includegraphics[width=\columnwidth, height=7cm]{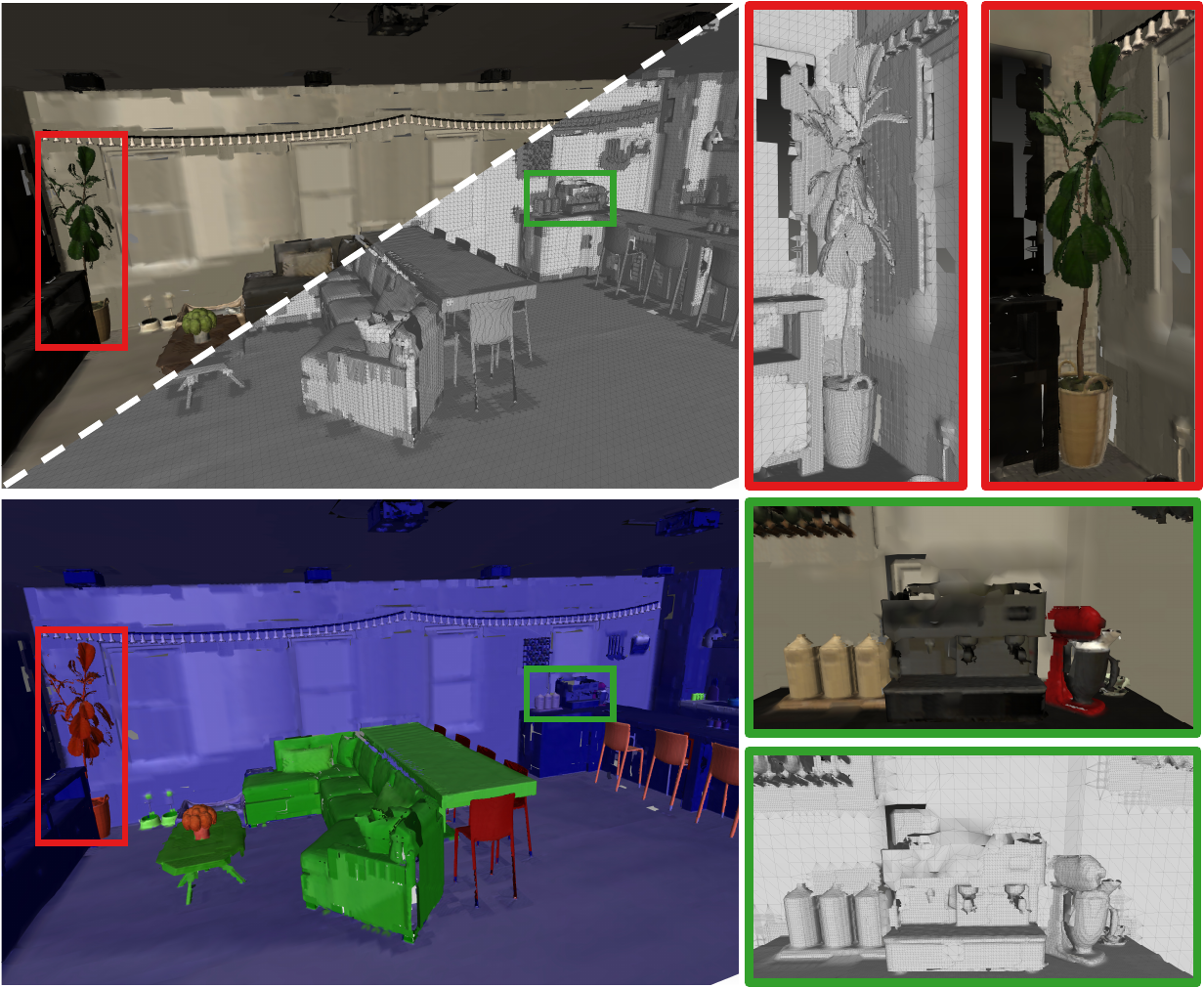}
    \caption{\textbf{\project{}}. Our method generates quality-adaptive semantic 3D maps of environments, where regions of different semantics and geometric complexity are reconstructed in different quality levels. An example map is shown here: 3D reconstructed mesh (top-left) and the semantic quality mask (bottom-left). Mask colors denote three quality levels, where \red{red is \textit{high}}, \green{green is \textit{middle}}, and \blue{blue is \textit{coarse}}. A plant reconstructed in high quality due to its semantic label is highlighted (top-right). Though the coffee machine based on its label should appear coarse, it is still mapped in fine resolution due to high geometric complexity (bottom-right).}
    \label{fig:teaser}
    \vspace{-15pt}
\end{figure}

Advancements in 3D sensing devices (\eg, Intel RealSense \cite{intelrealsense2023d435i}, Microsoft Kinect~\cite{microsoftkinect2023DK}, and Orbbec Astra~\cite{orbbec2023AstraSeries}) and semantic understanding \cite{kirillov2023segment,lai2023spherical,cai2021semantic} have enabled the reconstruction of an increasing number of semantic maps of environments in accuracy and detail. This is particularly useful for autonomous agents since they utilize such maps to perform tasks, \eg, navigation \cite{lin2018autonomous,aotani2017development} and object manipulation \cite{rusu2010semantic,breyer2021volumetric}. In recent practice, the common output of 3D reconstruction systems \cite{oleynikova2017voxblox,rosinol2020kimera,vespa2018efficient,pan2022voxfield} is a volumetric map of the environment that is uniform in the level of detail (single-resolution map). When the task requires a fine-grained and high-resolution reconstruction, \eg, for interacting with objects of small size or intricate geometry, the resulting map can lead to substantial computation and storage demands, which can be crucial for the operation of agents. 

We approach these shortcomings from the lens of not always needing `everything in anything', \ie, all information in the same level of detail, and address them by creating the 3D semantic maps in a quality-adaptive manner. 
Prior work has independently addressed building semantic maps \cite{PanopticFusion,SceneGraphFusion,Semantic:voxbloxplusplus,VoxbloxDiffusion,ModifiedVoxblox} and multi-resolution geometric mapping \cite{fuhrmann2011fusion,steinbrucker2014volumetric,stuckler2014multi,vespa2019adaptive,zienkiewicz2016monocular,kahler2015hierarchical,funk2021multi} to achieve accurate and memory-efficient reconstructions.
Except for \cite{schmid2022panoptic}, no other method has attempted to create quality-adaptive \textit{semantic} 3D maps. This method employs semantics to represent individual object instances in \textit{separate} 3D Truncated Signed Distance Field (TSDF) maps with different resolutions. However, since each map is created independently from the others and due to noisy semantic estimation, multiple maps may occupy the same spatial region without any mechanism to disambiguate across and merge them.

To address these limitations, we propose \textbf{\project{}}, a real-time method for quality-adaptive semantic 3D reconstruction with RGBD frames. 
Our main contribution is the \textit{first} adaptive semantic 3D mapping algorithm that generates directly a \textit{single} map with regions of different quality. 
In comparison to prior work on multi-resolution maps where the resolution is determined by the distance to the camera \cite{vespa2019adaptive, zienkiewicz2016monocular}, the quality per region is defined by the semantic label of an object and/or its geometric complexity. 
Our method is less computationally and storage demanding than single-resolution methods \cite{oleynikova2017voxblox} and it is faster and more accurate than the other semantic quality-adaptive method~\cite{schmid2022panoptic}. Hence, it has practical applicability to autonomous agents due to their limitations on computing, power, and storage. 

Given a lack of adaptive multi-resolution representations for semantic and 3D geometric data, we develop a new structure that can update reconstructed map regions and their quality level as new observations are received, building on top of an existing voxel hashing method~\cite{oleynikova2017voxblox}. We also propose a new approach to incrementally update the geometric complexity of the surface in each \textit{single} voxel. 
Furthermore, we estimate the camera pose and semantics of RGBD frames with a SLAM and a semantic segmentation method respectively, and use this information to build the map in an online manner. We propose a modified mesh generation method based on \cite{wald2020simple} to create a mesh from our multi-resolution map.
Last, we evaluate end-to-end the adaptive semantic reconstruction of \project{} and baselines in simulated and real-world environments using two state-of-the-art 3D semantic datasets. We will make the code and the data public. 
Our contributions are summarized as follows:
\begin{itemize}
    \item A real-time framework that generates a single quality-adaptive map, where areas that belong to different semantic groups and regions with intricate geometric details are distinctly reconstructed.
    \item A multi-resolution map representation that encapsulates geometric and semantic information and can be incrementally updated with newly acquired observations.
    \item An adaptive mesh generation approach that can handle voxels and their neighbors in different resolutions.
\end{itemize}

\section{Related Work}
\label{sec:rel_work}

\noindent
\textbf{Adaptive 3D Semantic Mapping.} 
Our focus is on methods that create real-time maps of the scene at different levels of quality. 
Prior work has mainly explored the creation of \textit{geometrically} adaptive maps. 
In \cite{fuhrmann2011fusion,steinbrucker2014volumetric}, 3D information is kept at multiple resolutions simultaneously. The coarse information is then used to regularize the fine resolution levels. 
This creates a large amount of redundant information, especially when considering large-scale 3D scenes. 
In \cite{stuckler2014multi}, the authors develop a SLAM approach that employs surfel-based 3D maps of incoming frames in different resolution levels, which are further associated per level to get the final 3D reconstruction of the scene. 
In \cite{zienkiewicz2016monocular}, the authors fuse depth frames into a multi-resolution triangular mesh that is adaptively tessellated based on the distance of the camera from the observed surface. 
Similarly, \cite{vespa2019adaptive} introduces an octree-based volumetric SLAM pipeline that integrates and renders depth images at an adaptive level of detail based on the camera distance. 
In \cite{kahler2015hierarchical}, the authors use a voxel-hashing approach to bypass the time-consuming traversal of tree structures and generate adaptive maps based on the geometric complexity of the surface. 
In our work, we address the problem of creating 3D semantic maps that adapt the geometry based on both geometric and semantic information. 
Our TSDF voxel-based formulation incorporates camera distance to define geometric and semantic accuracy.

In Panoptic Multi-TSDF~\cite{schmid2022panoptic}, similar to us, the authors use a TSDF voxel-based structure to acquire a semantic 3D map given RGBD frames. 
However, they represent each object instance in the scene in a separate TSDF voxel-based map that varies in terms of resolution depending on the semantic category of the object. 
Although this work handles semantic mapping with different resolutions, dividing the scene into multiple maps has certain limitations. 
Imperfect semantic segmentation and camera pose estimation can lead to duplicate reconstructions of spatial regions in these maps.
This occurs because semantic masks may overlap with adjacent categories when projected from 2D to 3D and individual maps are created in isolation without information exchange. This complicates merging the data into a single map due to the ambiguity in semantic interpretation. In contrast, we create a single map representation that handles regions of adaptive resolution as new data points are received and overcomes the above challenge because of the way it represents the scene.


\vspace{1mm}
\noindent
\textbf{3D Map Representations.} 
There exist multiple ways of representing 3D scenes, ranging from the use of 3D point clouds, to surfels \cite{whelan2015elasticfusion,mccormac2017semanticfusion}, voxels \cite{oleynikova2017voxblox,rosinol2020kimera}, 3D Gaussians \cite{kerbl3Dgaussians}, sparse representations \cite{nicholson2018quadricslam,yang2019cubeslam}, and neural implicit ones \cite{zhu2022nice,sucar2021imap}. 
For generating real-time maps that can operate on autonomous agents and allow them to perform other downstream tasks (\eg, navigation or object manipulation), voxel-based TSDF representations are commonly used. 
To further allow real-time generation, methods have focused on octrees \cite{hornung2013octomap} and voxel hashing \cite{niessner2013real,oleynikova2017voxblox,han2019fiesta}. In \cite{oleynikova2017voxblox}, voxel hashing was shown to be a more efficient method to query voxels compared to octrees \cite{hornung2013octomap}.
Hence, we build on the Voxblox \cite{oleynikova2017voxblox} voxel-hashing TSDF approach and contribute to it with a semantic adaptive structure and a fusion approach for generating and updating 3D semantic maps of adaptive resolution.


\begin{figure*}[tb]
    \centering
    \includegraphics[width=\textwidth]{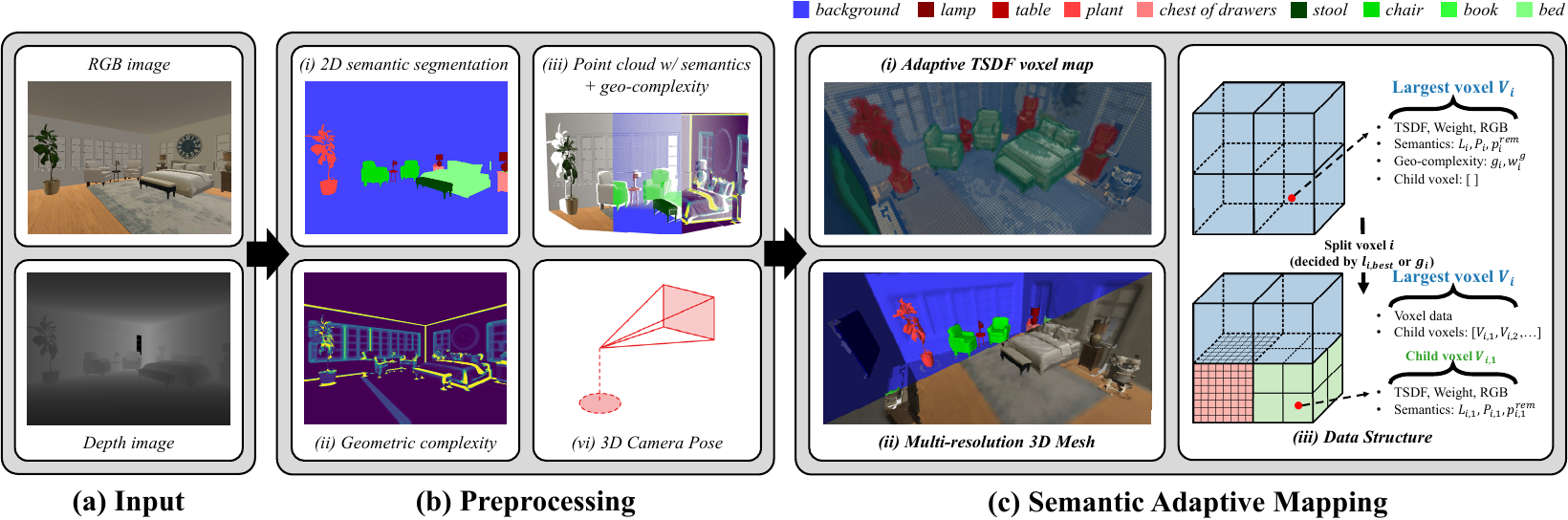}
    \caption{\textbf{Overview of \project{}.} (a) Given RGBD frames, we estimate (b-i) semantic segmentation and (b-iv) camera pose and compute (b-ii) geometric complexity. (c-i) We integrate geometric and semantic information (b-iii) on the TSDF voxel map. The geometric complexity and the semantic label will define the voxel size of that region of the map. (c-ii) shows the multi-resolution mesh output. The adaptive structure we use is shown in (c-iii).}
    \label{fig:pipeline}
    \vspace{-12pt}
\end{figure*}

\vspace{1mm}
\noindent
\textbf{Semantic SLAM.} 
Incorporating semantic information into SLAM-generated maps can be categorized into three types of methods: 
(i) \textit{Object detection-based}: Methods implement object-level detection (\eg, \cite{redmon2016you,liu2016ssd}) on RGB images to output 2D bounding boxes.
After further processing, they either use a parameterized way to represent the detected object, such as Quadrics~\cite{qian2021semantic} and the pose of a pre-modeled object \cite{salas2013slam++}, or further perform geometric segmentation on the depth map \cite{sunderhauf2017meaningful,hachiuma2019detectfusion}. 
(ii) \textit{Semantic segmentation-based}: Methods process semantic segmentation on 2D RGB images and build 3D geometric maps separately. 
The two outputs are fused with a Bayesian update to generate the semantic map \cite{mccormac2017semanticfusion,rosinol2020kimera}. 
(iii) \textit{Instance segmentation-based}: Such methods are similar to (ii). The main difference is that the RGB image is segmented to acquire object instances \cite{runz2018maskfusion,mccormac2018fusion++}.
One exception is the method of Grinvald et al.~\cite{grinvald2019volumetric}, which first segments a depth image and then utilizes the instance segmentation on an RGB image to refine the previous segments. 
We follow a semantic segmentation approach that is based on panoptic understanding \cite{mccormac2017semanticfusion,rosinol2020kimera}, but the proposed method can easily adapt to instances.

\vspace{1mm}
\noindent
\textbf{Mesh Generation.}
Marching Cubes \cite{lorensen1987marching} is widely used to extract mesh from a voxel-based map.
Although it is effective for fixed-size voxel maps \cite{oleynikova2017voxblox,rosinol2020kimera,niessner2013real}, modifications are required for multi-resolution ones.
To generate a mesh for a query voxel, a 2 $\times$ 2 $\times$ 2 cube is formed with its 7 neighbors. \cite{lorensen1987marching} requires the latter in the same size, which is not possible at the boundary of different resolutions.
\cite{vespa2019adaptive} proposes to use the coarsest resolution at the voxel boundary to ensure that all 8 voxels exist.
However, this ignores fine-level voxels. 
In contrast, we adapt the idea of \cite{wald2020simple} on iso-surface extraction to our specific data structure.
Such a method leverages information from the voxels at all levels.
Although \cite{funk2021multi} also claims to follow \cite{wald2020simple}'s approach for mesh generation from their multi-resolution voxel map, no explanation of the implementation is provided.
\section{\project{}}
\label{sec:method}

Given a set of RGBD frames, we use the RGB and depth images to estimate camera pose $C_k$ and predict semantic segmentation map $S_k$ using the RGB images only, where $k=1,2,...,K$ and $K$ is the total number of frames. We employ this information to create a quality-adaptive map in an online manner. 
Hereafter, $N \in \mathbb{N}^+$ is the total number of semantic labels that the semantic segmentation method can predict, $l$ is a semantic label from this set, and $l_{i,\text{best}}$ is the label with the highest probability in the voxel $V_i$.
Let us consider that the adaptive map has three resolution levels: fine, middle, and coarse.\footnote{Even though we describe the map assuming three levels of hierarchy, its depth in our implementation can be chosen arbitrarily, depending on the application at hand.} Each semantic label $l$ is associated with a level of the targeted reconstruction quality (\eg, fine) based on user preference. Per map region, the level of resolution is decided based on its semantic label and can also incorporate the geometric complexity of the observed surface.
Regarding the latter, thresholds are noted as $\theta_{r}$ where $r \in \{fine, middle, coarse\}$.
In the rest of this section, we describe the adaptive mapping process and map representation in detail. 
An overview of the pipeline is in Fig.~\ref{fig:pipeline}.

\vspace{1mm}
\noindent
\textbf{Adaptive Map Representation.} 
Our map representation, as in Voxblox~\cite{oleynikova2017voxblox}, uses a TSDF voxel grid $V$ to implicitly store geometric information, from which the 3D mesh of the mapped scene will be extracted with the use of Marching Cubes~\cite{lorensen1987marching}. This two forms of maps are shown as (c-i) and (c-ii) in Fig.~\ref{fig:pipeline}. 
In addition to the truncated distance, its weight, and color~\cite{oleynikova2017voxblox}, each voxel $V_i$ in our map stores its geometric complexity $g_i$, a weight $w^{g}_{i}$ representing the confidence in $g_i$, a vector $L_i$ of semantic labels that have been associated with this voxel, a vector $P_i$ of the probabilities corresponding to these semantics, and the probability $p^\text{rem}_{i}$ that corresponds to any non-associated semantic labels.
We assume a uniform probability distribution for all non-associated labels so that we can store their probabilities in a single scalar $p^\text{rem}_{i}$.
Each voxel is initialized with an empty vector for $L_i$ and $P_i$ and the probability $p^\text{rem}_{i}=1/N$.
As new RGB-D frames are processed, the probability of a semantic label $l$ may be updated for that voxel (see below for an explanation of the update process). 
If $l$ was previously associated with $V_i$, only its probability in $P_i$ is updated. 
Otherwise, $l$ and its probability will be added to the $L_i$ and $P_i$ vectors, respectively. 
Compared to allocating a single fixed-size vector per $V_i$ for the probabilities of all $N$ semantic classes, even if not associated with this voxel~\cite{rosinol2020kimera,mccormac2017semanticfusion}, our method uses less memory, especially when $N$ is large.  

So far, the described map representation is not adaptive. 
We introduce adaptivity by creating a hierarchy of parent-child voxels from the coarsest resolution (parent) to the finest (child). 
A given voxel $V_i$ in the voxel grid is initialized in the coarsest resolution when first created, \ie, it is initialized in the largest voxel size. 
If either the most likely semantic label $l_{i,\text{best}}$ of $V_i$ corresponds to a finer resolution level $r$ or the geometric complexity reaches the threshold $g_i \geq \theta_{r}$, this voxel will be subdivided by generating a vector of child voxels with the corresponding size of $r$. 
Furthermore, if $V_i$ already contains child voxels but both $l_{i,\text{best}}$ and $g_i$ get updated to one of the coarser resolutions, the child voxels will be removed from $V_i$ so that the voxel degrades back to a coarse representation. 
To avoid the loss of geometric information when the $l_{i,\text{best}}$ is uncertain, child voxels are removed only when $l_{i,\text{best}} \ge 0.95$. 
Please note that division and merging operations are defined based on the $l_{i,\text{best}}$ and $g_i$ of the voxel in the coarser resolution level.
This adaptive resolution structure is shown in Fig.~\ref{fig:pipeline} (c-iii).

  
\vspace{1mm}
\noindent
\textbf{Incorporating RGB-D Frames.}
With the depth map, RGB image, pose $C_{k}$, and semantic map $S_{k}$ at frame $k$, we create a semantically labeled 3D point cloud $PC_{k}$ in the world coordinate system (Fig.~\ref{fig:pipeline} (b-iii)). 
To avoid losing semantic information, especially when considering the noisy nature of predictions, instead of using a segmentation map that contains per pixel only the $l$ with the highest confidence score \cite{rosinol2020kimera}, we provide at most the four top-scoring semantic labels that have confidence score greater than the threshold $t = 0.1$. 
These semantic labels and their confidence scores are raycasted to voxels in $V$ per 3D point $pc_j$ in $PC_k$.
Similar to~\cite{oleynikova2017voxblox}, a ray that connects the camera center of frame $k$ with $pc_j$ is used to find those voxels whose absolute value of the truncated signed distance is smaller than their size.
This saves computational effort by only updating semantic information on voxels near the surface. 
We modify the raycasting in \cite{oleynikova2017voxblox} for adaptive resolution as described below.

\vspace{1mm}
\noindent
\textbf{Adaptive Raycasting.}
We use a modified version of the fast bundled raycasting in~\cite{oleynikova2017voxblox} but extend it to the resolution-adaptive setting. Before casting a ray on $V$, we need to decide which points from $PC_k$ may be redundant and hence can be skipped with a minimum loss of information. 
For a non-adaptive geometric map, a hash 3D grid with resolution $v_{grid} = \alpha v$, where $\alpha$ is a subsampling factor with default value 0.5 and $v$ is the voxel size of the TSDF map, keeps track of points in $PC_k$ that will be used to update $V$. 
Specifically, a point in $PC_k$ is discarded if the grid cell it falls into is already occupied by another 3D point originating from the same frame $k$. 
Since MAP-ADAPT has multiple resolutions (three in the described scenario), we initialize three grids with $v_{r,grid} = \alpha v_{r}$, where $v_{r}$ is the voxel size of quality level $r$ in the TSDF map. 
Each virtual grid is used to determine whether the point will be utilized to update voxels of the corresponding size.
Every point in $PC_k$ will be inserted into all three grids. If the position in $v_{r,grid}$ has already been occupied, this point will not be used to update voxels whose resolutions are level $r$. However, the same point might integrate information into voxels of another size $r'$ as long as the position in $v_{r',grid}$ is free.


\vspace{1mm}
\noindent
\textbf{Updating Voxel Probabilities.} Assume that $M_{j}$ is the set of four (or fewer) top-scoring semantic labels for point $pc_j$ and $P_{j}(l|S_{k})$ is the probability of the label of point $pc_j$ to be semantic label $l$.
When assigning semantic information from $pc_j$ in $PC_k$ to a voxel $V_i$, we use the probabilities that are already associated with $pc_j$ for the top-scoring semantic labels in $M_{j}$. 
For all other labels, we assume a uniform probability distribution.
To avoid exceedingly fast convergence to a specific label for $V_i$, we empirically define a lower bound $\xi = 0.01$. Specifically, $\forall l \notin M_{j}$, its probability is given by:
\begin{equation}
    P_{j}(l \; | \; S_{k}) = 
    \max \left(\xi,\frac{1-\sum_{m \in M_{j}}P_{j}(l_m \; | \; S_{k})}{N- \text{sizeof}(M_{j})} \right).
\end{equation}
Similar to~\cite{mccormac2017semanticfusion,rosinol2020kimera}, when new frames are incorporated in $V$, a Bayesian update is utilized to update the semantic probabilities of voxel $V_i$. 
Given the 3D point $pc_{j}$, the probability of a voxel $V_i$ to be semantic label $l$ after $k$ frames $P_{i}(l \; | \; S_{1,...,k})$ is updated by the following rules:
\begin{equation}
    P_{i}(l \; | \; S_{1,...,k}) = \frac{1}{Z}P_{i} \left(l \; | \; S_{1,...,k-1} \right) {[P_{j}(l \; | \; S_{k})]}^{{w}_{j}},
\end{equation}
where $Z$ is a normalization term for the probabilities so that they will sum to 1 and $w_{j}= 1 / z_{j}^{2}$ 
is a weight function that depends on the depth measurement $z_{j}$ of point $pc_{j}$ in the depth frame $k$.

\vspace{1mm}
\noindent
\textbf{Estimating Geometric Complexity.} 
As certain tasks require increased precision in understanding the geometric details of objects beyond semantic distinctions, we employ a voxel-wise geometric complexity measurement to determine the reconstruction quality level.
This involves assessing the change of curvature~\cite{pauly2002efficient,rusu2010semantic} on the projected points at each frame $k$ and incrementally updating this value in $V$. 
For a point $pc_{j}$ in $PC_{k}$, the eigenvalues $\lambda^{j}_{1},\lambda^{j}_{2},\lambda^{j}_{3}$ of the respective 3D structure tensor~\cite{jutzi2009nearest} are extracted, with $\lambda^{j}_{1}\geq\lambda^{j}_{2}\geq\lambda^{j}_{3}\geq0$. The change of curvature at this point $CC_{j}$ is calculated as 
$CC_{j} = \lambda^{j}_{3} / (\lambda^{j}_{1}+\lambda^{j}_{2}+\lambda^{j}_{3})$.
For a voxel $V_{i}$ passed by the ray of $pc_{j}$, its geometric complexity $g_{i}$ and weight $w_{i}^{g}$ are updated as follows:
\begin{equation}
\begin{aligned}
    g_{i} &\leftarrow \frac{w^{g}_{i}g_{i}+w_{j}CC_{j}}{w^{g}_{i} + w_{j}} \\
    w^{g}_{i} &\leftarrow \min{(w^{g}_{i} + w_{j}, W_{max})}
    \end{aligned}
\end{equation}
$W_{max}$ is the same upper bound as in updating TSDF values.

\vspace{1mm}
\noindent
\textbf{Multi-resolution Mesh Generation.}
We generate the final 3D mesh with Marching Cubes \cite{lorensen1987marching} in a bottom-up fashion. When generating the mesh, we traverse all coarse-resolution voxels $V_{coarse}$. If a voxel in $V_{coarse}$ has children -- \ie is split into a finer resolution, the mesh will be generated on its child voxels. To mesh a voxel with \cite{lorensen1987marching}, we need the TSDF values and coordinates of its 7 neighbors to form a cube. However, \cite{lorensen1987marching} requires all 8 voxels to be in the same resolution, which is not always feasible in our multi-resolution map.

\begin{figure}[tb]
    \centering
    \begin{subfigure}[b]{0.45\columnwidth}
        \centering
        \includegraphics[width=0.8\columnwidth]{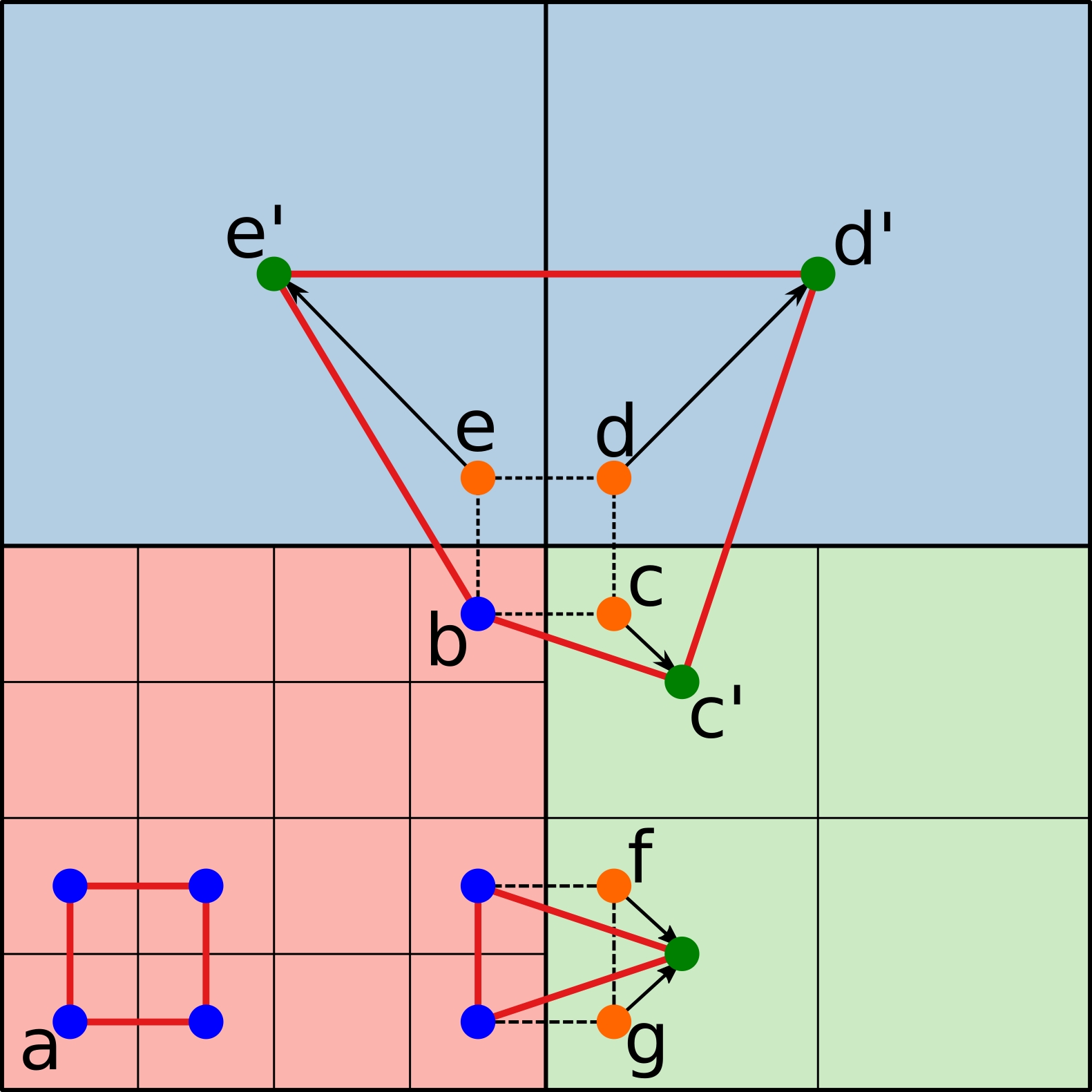}
        \caption{Mesh generation.}
        \label{fig:mesh_generation}
    \end{subfigure}
    \hfill
    \begin{subfigure}[b]{0.5\columnwidth}
        \centering
        \includegraphics[width=0.95\columnwidth]{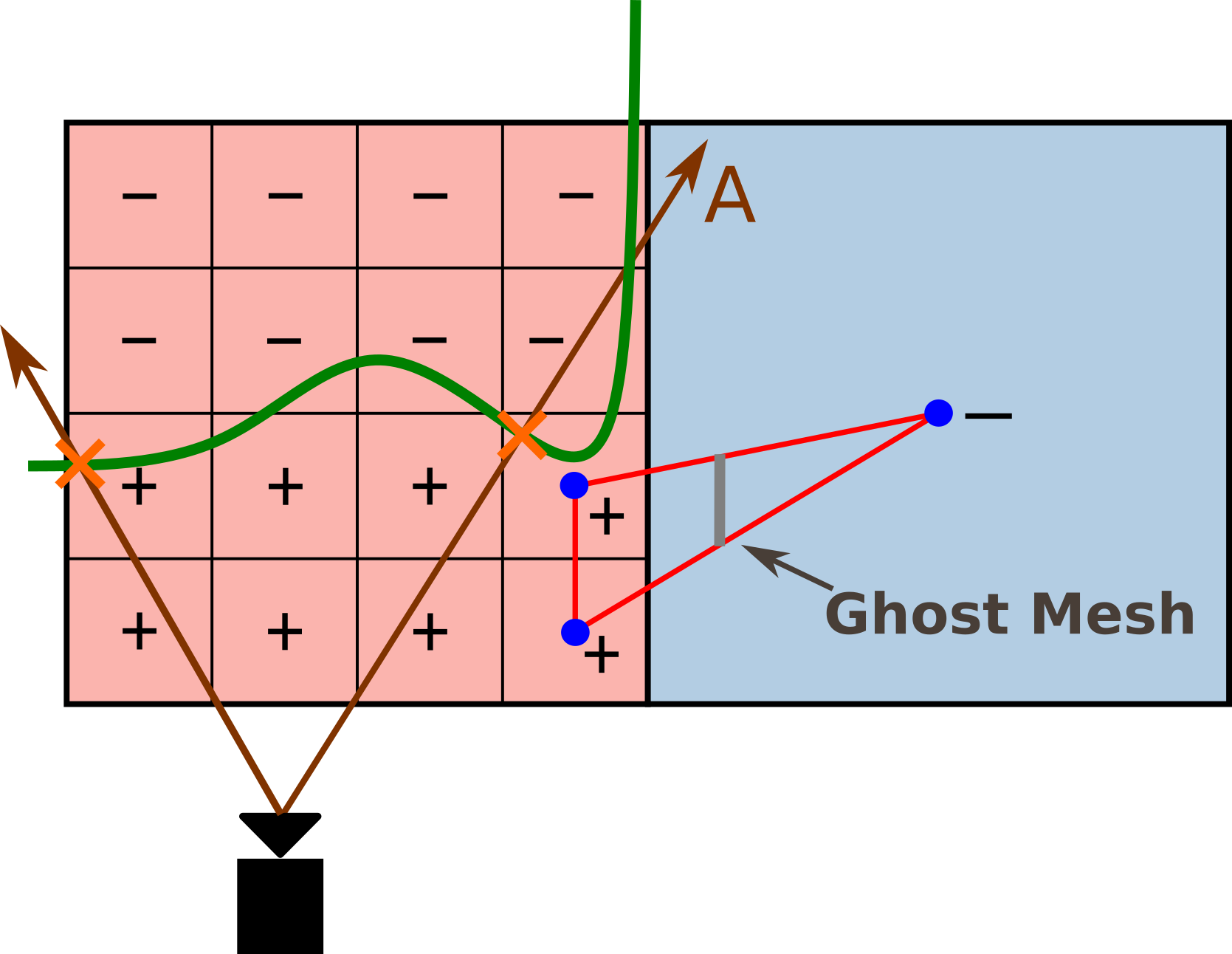}
        \caption{An example of ghost mesh.}
        \label{fig:ghost_mesh}
    \end{subfigure}
    \caption{\textbf{Illustration of forming a cube to generate a mesh from our multi-resolution map.} (a) When a neighboring voxel of the queried resolution (\textit{\textbf{\textcolor{eccvorange}{orange node}}}) does not exist, the corresponding coarser-resolution one (\textit{\textbf{\textcolor{eccvgreen}{green node}}}) will be used instead. (b) A ghost mesh is generated at the boundary of resolutions.}
    \label{fig:mesh_both}
    \vspace{-15pt}
\end{figure}

To construct the 8-voxel cube, we initiate the process with voxels at the finest resolution. If any of the 8 voxels is absent at this level, it is substituted by its corresponding voxel at a coarser resolution. This process is illustrated in Figure~\ref{fig:mesh_generation} using a 2D grid for simplification. Voxel \textbf{\textit{a}} shows the typical mesh generation approach in fixed-size maps, where \textbf{\textit{a}} and its neighbors all belong to the finest resolution.
When attempting to mesh \textbf{\textit{b}}, which is at the finest level, its neighbors (\textbf{\textit{c}}, \textbf{\textit{d}}, and \textbf{\textit{e}}) are not available there. Consequently, we substitute them with their coarse counterparts \textbf{\textit{c'}}, \textbf{\textit{d'}}, and \textbf{\textit{e'}}.
This substitution may result in the formation of different geometric structures, such as triangles or lines, instead of hexahedra.
For instance, when multiple fine-resolution voxels like \textit{\textbf{f}} and \textit{\textbf{g}} are substituted by the same coarse-resolution voxel, it leads to collapsed edges where endpoints coincide.
As noted in~\cite{wald2020simple}, \cite{lorensen1987marching} can still process these geometries effectively as if they were regular hexahedra; no mesh is generated along the collapsed edges since both endpoints have the same TSDF value.


The other challenging issue is that non-existent meshes (ghost meshes) may be generated near the surface of objects that occupy voxels in finer resolution. This primarily occurs because the adjacent voxels in free space are in the coarsest resolution, leading to reduced accuracy in their TSDF values.
An example is in Figure~\ref{fig:ghost_mesh}, where we split a voxel to the finest resolution because it contains a surface with high geometric complexity, while its right neighbor remains in the coarse resolution. 
When ray \textbf{\textit{A}} is integrated into the map, the blue voxel, which is supposed to be empty, will also be updated since the ray passes through a small part of it. As a result, the voxel will be assigned a negative TSDF value. Since two of its neighbors have a positive TSDF value, a ghost mesh will be generated there. To mitigate this problem, when a voxel is split to a finer resolution, we also split all neighboring voxels to the same one. 
Though it will lead to higher quality reconstruction on regions which should have coarser resolution, it significantly improves the quality of the generated mesh for fine-level semantics. 



\section{Experiments}
\label{sec:exps}

We evaluate \project{'s} performance on creating accurate and complete geometric and semantic 3D maps with adaptive resolution, and compare with the fixed voxel size Voxblox \cite{oleynikova2017voxblox} at different resolution levels, as well as with Panoptic Multi-TSDFs \cite{schmid2022panoptic}.
We choose the following three levels of quality (voxel size): fine (1 cm), middle (4 cm), and coarse (8 cm). We use all three in the adaptive methods (ours and \cite{schmid2022panoptic}), whereas for the fixed-size one, we compare to three different instantiations of it, one per resolution.
Results from two versions of \project{} are provided; \textbf{\project{-S}} decides to divide a voxel only based on its semantic label, whereas \textbf{\project{-SG}} decides based on semantic label and/or geometric complexity.

We report results on the Habitat Synthetic Scene Dataset (HSSD) \cite{khanna2023hssd} and on the real-world ScanNet \cite{dai2017scannet} dataset.
The threshold of geometric complexity is chosen as $\theta_{middle} = 0.05, \theta_{fine} = 0.1$.
Since the motivation of the system is task-driven, giving users the freedom to choose which categories to reconstruct in fine quality and which are unimportant, in our experiments we randomly allocate semantic categories per level of quality and we repeat this 5 times; results are averaged over them.
For HSSD, we randomly assign the 28 semantic categories provided into the three levels of quality.
For the 40 NYUv2 \cite{silberman2012indoor} labels used in ScanNet, we allocate those corresponding to the HSSD categories to the same quality level and randomly assign the rest. In the supplementary material, we provide results with allocating semantics per quality level by their physical size.


\begin{table*}[tb]
    \footnotesize
    \centering
    \resizebox{2\columnwidth}{!}{
    \begin{tabular}{c|cc|ccccc}
        \toprule
          & \textbf{Method} & \makecell{\textbf{Reconstruction} \\ \textbf{Quality} (cm)} & \makecell{\textbf{Completion} \\ \textbf{Error} (cm) $\downarrow$} & \makecell{\textbf{Compl. $<$5cm} \\ \textbf{Ratio} (\%) $\uparrow$} & \makecell{\textbf{Geometric} \\ \textbf{Error} (cm) $\downarrow$} & \makecell{\textbf{Semantic} \\ \textbf{Accuracy} (\%) $\uparrow$} & \makecell{\textbf{Semantic} \\ \textbf{mIoU} (\%) $\uparrow$} \\ 
         \midrule \midrule

         \parbox[t]{2mm}{\multirow{4}{*}{\rotatebox[origin=c]{90}{\textit{\red{@1cm}}}}}
         & Voxblox \cite{oleynikova2017voxblox} (fixed) & Fine [1] & \textbf{\red{2.49 $\pm$ 2.80}} & \textbf{\red{88.74}} & \red{4.14 $\pm$ 4.49} & \red{12.96} & \red{6.62} \\
         & Multi-TSDFs \cite{schmid2022panoptic} & Multi-level [1-4-8] & \red{2.74 $\pm$ 4.00} & \red{85.59} & \textbf{\red{4.10 $\pm$ 6.53}} & \phantom{1}\red{8.58} & \red{4.86} \\
         & \textbf{\project{-S}} & \textbf{Adaptive [1-4-8]} & \red{2.54 $\pm$ 2.92} & \red{88.15} & \red{4.18 $\pm$ 4.62} & \textbf{\red{13.12}} & \textbf{\red{6.74}} \\
         & \textbf{\project{-SG}} & \textbf{Adaptive [1-4-8]} & \red{2.53 $\pm$ 2.84} & \red{88.34} & \red{4.19 $\pm$ 4.57} & \textbf{\red{13.12}} & \textbf{\red{6.74}} \\
        \midrule 
        \parbox[t]{2mm}{\multirow{4}{*}{\rotatebox[origin=c]{90}{\green{\textit{@4cm}}}}}
         & Voxblox \cite{oleynikova2017voxblox} (fixed) & Middle [4] & \green{3.06 $\pm$ 3.50} & \green{84.39} & \green{4.10 $\pm$ 4.16} & \textbf{\green{40.00}} & \green{16.01} \\
         & Multi-TSDFs \cite{schmid2022panoptic} & Multi-level [1-4-8] & \green{3.09 $\pm$ 3.83} & \green{83.29} & \green{4.18 $\pm$ 6.46} & \green{10.57} & \phantom{1}\green{6.41} \\
         & \textbf{\project{-S}} & \textbf{Adaptive [1-4-8]} & \green{2.89 $\pm$ 3.45} & \green{86.12} & \green{4.05 $\pm$ 4.19} & \green{39.69} & \textbf{\green{16.26}} \\
         & \textbf{\project{-SG}} & \textbf{Adaptive [1-4-8]} & \textbf{\green{2.67 $\pm$ 3.25}} & \textbf{\green{88.04}} & \textbf{\green{3.85 $\pm$ 4.13}} & \green{39.88} & \green{16.21} \\
        \midrule 
        \parbox[t]{2mm}{\multirow{4}{*}{\rotatebox[origin=c]{90}{\textit{\blue{@8cm}}}}} 
         & Voxblox \cite{oleynikova2017voxblox} (fixed) & Coarse [8] & \blue{3.59 $\pm$ 3.59} & \blue{77.93} & \blue{4.57 $\pm$ 6.11} & \textbf{\blue{60.38}} & \textbf{\blue{21.46}} \\ 
         & Multi-TSDFs \cite{schmid2022panoptic} & Multi-level [1-4-8] & \blue{3.42 $\pm$ 3.79} & \blue{79.86} & \textbf{\blue{4.05 $\pm$ 5.89}} & \blue{49.59} & \phantom{1}\blue{8.85} \\
         & \textbf{\project{-S}} & \textbf{Adaptive [1-4-8]} & \blue{3.43 $\pm$ 3.47} & \blue{79.94} & \blue{4.53 $\pm$ 5.95} & \textbf{\blue{60.38}} & \blue{21.18} \\
         & \textbf{\project{-SG}} & \textbf{Adaptive [1-4-8]} & \textbf{\blue{3.10 $\pm$ 3.27}} & \textbf{\blue{83.56}} & \blue{4.53 $\pm$ 5.89} & \textbf{\blue{60.38}} & \blue{21.17} \\
        \bottomrule
    \end{tabular}
    }
    \caption{\textbf{Evaluation per quality level on HSSD \cite{khanna2023hssd}.} @XXcm represents the evaluation on the regions of semantics corresponding to the resolution level of XX (cm). Best values per evaluation level are in \textbf{bold}.}
    \label{tab:map_resolution}
    \vspace{-8pt}
\end{table*}

We employ the commonly used ORB-SLAM2 \cite{mur2017orb} as the visual SLAM module for its robust and real-time behavior; any other SLAM approach could also be used. 
We employ the Light-weight Refinenet \cite{nekrasov2018light} as the segmentation module\footnote{Even though we demonstrate \project{} with object categories, other semantic information can be used, \eg, material, function, change.}, for allowing real-time processing while providing good segmentation results on unseen data. 
Similarly, other segmentation methods could be used, especially if processing time is not a concern. 
We sample training and validation data from the 125 HSSD scenes in the train split to train a Lightweight RefineNet model for our experiments on this dataset. Provided by~\cite{nekrasov2018light}, a pre-trained model on the NYUv2 dataset is used for ScanNet.

\vspace{1mm} \noindent \textbf{Metrics.} For geometric evaluation, we report: (i) completion error (cm), \ie, the mean Euclidean distance of all ground truth (GT) 3D points from the closest reconstructed ones; (ii) completion ratio for all GT points that have less than 5 cm distance from the closest reconstructed point; and (iii) geometric error (cm), \ie, the mean Euclidean distance of all reconstructed points from the closest GT ones. The reconstructed 3D points are sampled from the generated mesh. The GT points are the aggregated projections from all depth frames, using GT camera pose. The geometric metrics are calculated separately for 3 different quality levels. Each GT point will be classified as 1cm, 4cm, or 8cm based on its GT semantic label. For each point $P_i$ sampled from the reconstructed mesh, we identify the nearest point $P_{gt}$ in the GT map.
We then evaluate $P_{i}$ based on the level corresponding to the semantic label of that closest $P_{gt}$ regardless of the predicted semantic label of $P_{i}$.
For each of the three semantic levels, evaluation is between the sampled points and GT points based on the latter's quality level. For semantic evaluation, we follow standard approaches~\cite{rosinol2020kimera} and report the overall portion of correctly labeled voxels (Accuracy) and the mean Intersection over Union (mIoU). We report map size in megabytes (MB) and runtime in milliseconds (ms).

\vspace{1mm} \noindent \textbf{Experimental Setup.} All experiments are performed on an AMD Ryzen 7 5800H CPU. The only component in our system that requires GPU is the 2D semantic segmentation, which takes 37ms per frame on a GeForce RTX 2080 GPU for Light-weight Refinenet~\cite{nekrasov2018light}.

\vspace{1mm} \noindent \textbf{HSSD Dataset.} 
The HSSD dataset \cite{khanna2023hssd} consists of high-quality 3D scenes on the scale of an entire residence with fully human-authored 3D interiors.
To generate sequences of frames for SLAM-based reconstruction, the dataset is commonly used within the Habitat \cite{szot2021habitat,habitat19iccv} simulation environment, which can render RGBD frames from the underlying 3D model given arbitrary 3D camera poses. 
We manually record camera trajectories in the scenes and use the rendered RGBD frames in our experiments.
We develop our method on the training scenes and evaluate on the \textit{open} validation scenes without parameter tuning.
We create 43 subscenes from the validation split and ensure they contain at least one semantic category per quality level in each subscene. Statistics on these scenes are in the suppl. material.


Results on geometric and semantic evaluation are shown in Table~\ref{tab:map_resolution}. We employ colors to differentiate the evaluation of regions from different quality levels and report only directly comparable methods; \eg, at \blue{\textit{Eval.\ @8cm}}, \project{} and \cite{schmid2022panoptic} are directly comparable with the Voxblox fixed-size on 8cm. Results for the fixed-size methods on other resolutions are included in the supplementary material. We compute metrics per quality level \textit{only} based on the GT semantic regions that correspond to this level.

As shown in Table~\ref{tab:map_resolution}, both versions of \project{} achieve performance similar to fixed (1cm) reconstruction in regions where semantics are at the finest level.
In regions of semantics belonging to middle and coarse quality, \project{-S} performs slightly better than the corresponding fixed size \cite{oleynikova2017voxblox} since we split the neighboring voxels of fine-quality semantic voxels, thus these regions will be closer to ground truth points.
\project{-SG} outperforms all methods in terms of completion error in the middle and coarse semantic regions as it generates finer resolution voxels in regions with high geometric complexity even if their semantics are not allocated in the fine-quality set.
The geometric error of \project{-SG} does not show an advantage and is even higher than \cite{schmid2022panoptic} in the coarsest regions due to errors in the estimated camera pose.
A detailed reconstruction will lead to even higher geometric error if it is reconstructed in the wrong position (see Section~\ref{sec:ablation} for GT camera pose results).
For the map reconstructed using the fixed size method \cite{oleynikova2017voxblox} with the finest quality (1 cm), the geometric error in the \textit{coarse} semantic regions is also large ($5.11\pm 6.37 cm$).
This result is reported in the supplementary material.
In contrast, \cite{schmid2022panoptic} generates a relatively incomplete reconstruction (higher completion error) in these challenging regions. This means there are fewer points from which to compute the geometric error, which partially explains the lower values.
The other issue of \cite{schmid2022panoptic} is that it generates overlapping mesh regions across the individual semantic maps due to the noisy semantic estimation, as explained in the related work. 
As a result, \cite{schmid2022panoptic} performs significantly worse than \project{} in all semantic evaluation metrics.



\begin{figure*}
    \centering
    \includegraphics[width=0.99\textwidth]{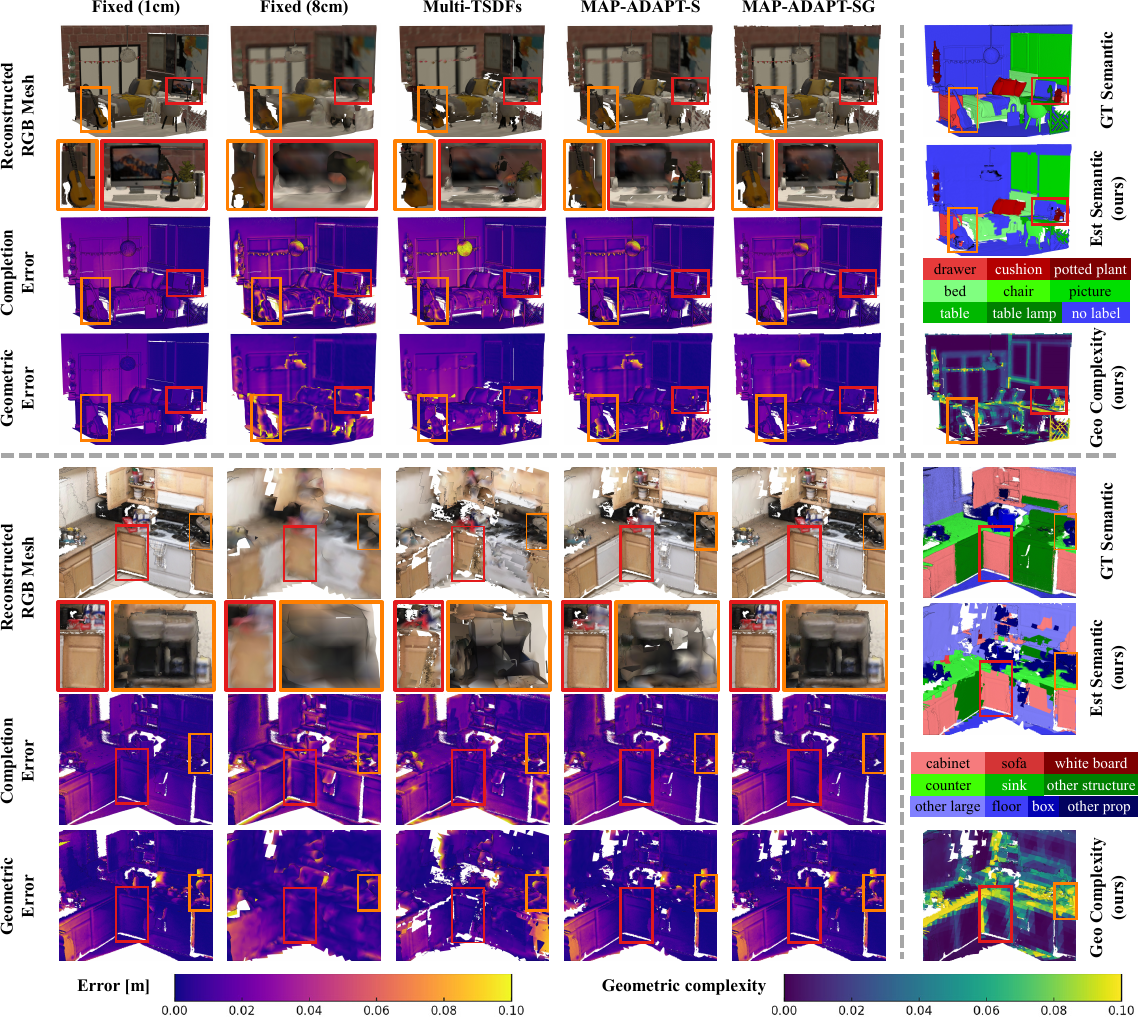}
    \caption{\textbf{Reconstruction results per method}. Top example is on HSSD and bottom one on ScanNet datasets. Geometric and completion errors are shown as heatmaps; the darker the color, the closer to the GT geometry. For semantic map, results are colorized per quality level; different semantics in the same quality level range from brighter to darker. Another heatmap is used to show the estimated geometric complexity. We highlight regions that are classified into high-quality semantics (\textit{\red{red block}}) or have large geometric variance (\textit{\textcolor{eccvorange}{orange block}}). \textit{Best viewed on screen.}}
    \label{fig:heatmap}
    \vspace{-10pt}
\end{figure*}

An example of the generated map is shown in Figure~\ref{fig:heatmap} (top). 
The fixed-size Voxblox versions provide less accurate geometry on the overall map as the quality level goes from fine to coarse. 
This is visible with the increasingly brighter colored regions for completion and geometric errors. 
For Muti-TSDFs \cite{schmid2022panoptic} and \project{}, we can observe the adaptive reconstruction from the various errors per semantic quality level. 
The red regions (fine quality) in the GT semantic map have darker colors in the visualizations of completion and geometric error, and the blue regions (coarse quality) are brighter in the error map.
Comparing ours with \cite{schmid2022panoptic}, the completion error maps of \project{-S} and \project{-SG} are darker throughout the scene.
On the geometric error map, we generally have a better result. 
However, objects with high error (\eg chandelier) are not reconstructed in \cite{schmid2022panoptic}, explaining why it has less geometric error on average, as stated above.
In the geometric complexity color map, we observe that \project{-SG} manages to capture regions having rich geometric information and those regions have much less error compared to \project{-S}.
We highlighted regions with high geometric complexity (guitar) and with high-quality semantics (plant), where \project{-SG} generates the most detailed and sharp reconstruction.

We also report the map size and runtime per method in Table~\ref{tab:runtime}. 
Compared to Voxblox (1cm), \project{-S} occupies 3.5 times less memory and is also faster in updating the TSDF values.
\project{-SG} needs more storage and time since it reconstructs more high-quality regions but still consumes less than Voxblox (1cm). In contrast, \cite{schmid2022panoptic} takes substantially more time to perform TSDF updates since \cite{schmid2022panoptic} generates multiple TSDF maps per instance and requires an additional process to track. Both versions of \project{} need more time to generate the mesh due to the complex generation of meshes on the border of different voxel resolutions. However, mesh generation is only executed once at the end of reconstruction. We provide statistics and analysis on voxel percentage per quality level in the supplementary material.

\begin{table}[htbp!]
    \centering
    \resizebox{0.95\columnwidth}{!}{
    \begin{tabular}{cc|ccc}
        \toprule
        \multirow{2}{*}{\textbf{Method}} & \multirow{2}{*}{\makecell{\textbf{Map Size} \\ (MB) $\downarrow$}} & \multicolumn{2}{c}{\textbf{Runtime} (ms) $\downarrow$} \\
         & & Update TSDF & Generate Mesh \\
        \midrule \midrule
        Voxblox~\cite{oleynikova2017voxblox} (1cm) & 1225.30 & \phantom{1}89.61 $\pm$ \phantom{1}14.16 & \phantom{1}631.66 $\pm$ 307.50 \\
        Voxblox~\cite{oleynikova2017voxblox} (4cm) & \phantom{1}\phantom{1}60.39 & \phantom{1}46.02 $\pm$ \phantom{1}\phantom{1}7.47 & \phantom{1}\phantom{1}31.76 $\pm$ \phantom{1}11.15 \\
        Voxblox~\cite{oleynikova2017voxblox} (8cm) & \phantom{1}\phantom{1}\textbf{12.53} & \phantom{1}\textbf{39.86 $\pm$ \phantom{1}\phantom{1}6.40} & \textbf{\phantom{1}\phantom{1}\phantom{1}8.83 $\pm$ \phantom{1}\phantom{1}2.90}\\
        \midrule
        Multi-TSDFs \cite{schmid2022panoptic} & \phantom{1}266.91 & 201.45 $\pm$ 212.02 & \phantom{1}\underline{\textbf{203.81 $\pm$ 155.16}} \\        
        \textbf{\project{-S}} & \phantom{1}\underline{\textbf{265.21}} & \phantom{1}\underline{\textbf{54.62 $\pm$ \phantom{1}10.99}} & \phantom{1}638.55 $\pm$ 384.82 \\
        \textbf{\project{-SG}} & \phantom{1}469.85 & \phantom{1}71.79 $\pm$ \phantom{1}11.87 & 1252.54 $\pm$ 606.16 \\
        \bottomrule
    \end{tabular}
    }
    \caption{\textbf{Evaluation on map size and runtime.} \textit{Best} values are \textbf{bold}. \textit{Best} of multi-resolution methods are in \underline{\textbf{underlined bold}}. Note that update TSDF is processed at each frame, whereas mesh generation only needs to be executed once at the end.}
    \label{tab:runtime}
    \vspace{-12pt}
\end{table}

\noindent \textbf{ScanNet Dataset.} To understand the behavior of \project{} given real-world RGBD frames, we evaluate on the ScanNet dataset \cite{dai2017scannet}.
It consists of 3D scenes on the scale of a room and includes 3D mesh reconstructions, as well as the sequences of RGBD frames that were used for the reconstruction.
We evaluate our approach on 38 randomly selected scenes from the \textit{open} validation set.\footnote{
We employ the validation set since the test set does not have publicly available annotations.}
The results are in Table~\ref{tab:scannet}. We can observe that all methods perform less well on this real-world dataset, given the blurriness in the frames and noisy sensors. Despite this, \project{} achieves comparable results to fix-size (1cm) on fine-quality regions and performs better in semantic accuracy and completion error on coarser regions. Results of \cite{schmid2022panoptic} are similar to the HSSD dataset. An example of the generated maps is in Figure~\ref{fig:heatmap} (bottom). Fixed-size Voxblox has a similar behavior as on HSSD, and so does ours -- \eg, error maps are comparable per quality level. \project{-S} and \project{-SG} still provide lower completion error over all quality levels vs. \cite{schmid2022panoptic}.
The map reconstructed by \cite{schmid2022panoptic} exhibits a more irregular structure with more holes in the cabinet and several ghost meshes, indicating that it is more affected by the noisy pose estimation and depth data.

\begin{table}[htbp!]
    \centering
    \resizebox{\columnwidth}{!}{
    \begin{tabular}{c|cc|ccccc}
        \toprule
        & \textbf{Method} & \makecell{\textbf{Reconstruction} \\ \textbf{Quality} (cm)} & \makecell{\textbf{Completion} \\ \textbf{Error} (cm) $\downarrow$} & \makecell{\textbf{Compl. $<$5cm} \\ \textbf{Ratio} (\%) $\uparrow$} & \makecell{\textbf{Geometric} \\ \textbf{Error} (cm) $\downarrow$} & \makecell{\textbf{Semantic} \\ \textbf{Accuracy} (\%) $\uparrow$} & \makecell{\textbf{Semantic} \\ \textbf{mIoU} (\%) $\uparrow$} \\ 
        \midrule 
        \parbox[t]{2mm}{\multirow{4}{*}{\rotatebox[origin=c]{90}{\textit{\red{@1cm}}}}}
        & Voxblox \cite{oleynikova2017voxblox} (fixed) & Fine [1]  & \textbf{\red{3.21 $\pm$ 4.92}} & \textbf{\red{82.61}} & \phantom{1}\red{7.08 $\pm$ 13.38} & \red{10.36} & \textbf{\red{6.60}} \\
        & Multi-TSDFs \cite{schmid2022panoptic} & Multi-level [1-4-8] & \red{3.75 $\pm$ 5.69} & \red{77.42} & \phantom{1}\textbf{\red{5.53 $\pm$ 10.47}} & \phantom{1}\red{6.55} & \red{4.41} \\
        & \textbf{\project{-S}} & \textbf{Adaptive [1-4-8]} & \red{3.36 $\pm$ 5.20} & \red{81.57} & \phantom{1}\red{6.31 $\pm$ 11.51} & \textbf{\red{10.40}} & \textbf{\red{6.60}} \\
        & \textbf{\project{-SG}} & \textbf{Adaptive [1-4-8]} & \red{3.27 $\pm$ 5.03} & \red{82.27} & \phantom{1}\red{6.84 $\pm$ 12.99} & \red{10.36} & \red{6.59} \\
        \midrule 
        \parbox[t]{2mm}{\multirow{4}{*}{\rotatebox[origin=c]{90}{\green{\textit{@4cm}}}}}
        & Voxblox \cite{oleynikova2017voxblox} (fixed) & Middle [4] & \green{4.93 $\pm$ 6.52} & \green{69.52} & \phantom{1}\green{7.90 $\pm$ 14.80} & \phantom{1}\textbf{\green{9.07}} & \textbf{\green{5.76}} \\
        & Multi-TSDFs \cite{schmid2022panoptic} & Multi-level [1-4-8] & \green{4.43 $\pm$ 6.80} & \green{74.94} & \phantom{1}\textbf{\green{6.95 $\pm$ 13.28}} & \phantom{1}\green{4.47} & \green{3.23} \\
        & \textbf{\project{-S}} & \textbf{Adaptive [1-4-8]}  & \green{4.24 $\pm$ 6.22} & \green{75.62} & \phantom{1}\green{8.02 $\pm$ 14.84} & \phantom{1}\green{8.89} & \green{5.71} \\
        & \textbf{\project{-SG}} & \textbf{Adaptive [1-4-8]}& \textbf{\green{3.91 $\pm$ 6.02}} & \textbf{\green{78.82}} & \phantom{1}\green{8.58 $\pm$ 16.20} & \phantom{1}\green{9.00} & \green{5.73} \\
        \midrule 
        \parbox[t]{2mm}{\multirow{4}{*}{\rotatebox[origin=c]{90}{\textit{\blue{@8cm}}}}} 
        & Voxblox \cite{oleynikova2017voxblox} (fixed) & Coarse [8] & \blue{6.48 $\pm$ 7.30} & \blue{55.36} & \blue{11.43 $\pm$ 17.92} & \textbf{\blue{19.05}} & \textbf{\blue{8.94}} \\
        & Multi-TSDFs \cite{schmid2022panoptic} & Multi-level [1-4-8] & \blue{5.23 $\pm$ 7.02} & \blue{67.00} & \phantom{1}\textbf{\blue{9.02 $\pm$ 15.02}} & \blue{14.10} & \blue{5.28} \\
        & \textbf{\project{-S}} & \textbf{Adaptive [1-4-8]} & \blue{5.01 $\pm$ 6.64} & \blue{67.77} & \phantom{1}\blue{9.94 $\pm$ 16.07} & \textbf{\blue{19.05}} & \textbf{\blue{8.94}} \\
        & \textbf{\project{-SG}} & \textbf{Adaptive [1-4-8]} & \textbf{\blue{4.27 $\pm$ 6.19}} & \textbf{\blue{74.51}} & \phantom{1}\blue{9.48 $\pm$ 15.82} & \textbf{\blue{19.05}} & \textbf{\blue{8.94}} \\
        \bottomrule
    \end{tabular}
    }
    \caption{\textbf{Evaluation per quality level on Scannet \cite{dai2017scannet}.} @XXcm represents the evaluation on the regions of semantics corresponding to the resolution level of XX (cm). Best values per evaluation level are in \textbf{bold}.}
    \label{tab:scannet}
    \vspace{-10pt}
\end{table}

\subsection{Ablation Studies}
\label{sec:ablation}
In this section, to further evaluate our design choices, we provide additional experiments on all 43 scenes from the HSSD dataset with 1 random semantic quality allocation.

\vspace{1mm}
\noindent \textbf{GT pose and semantics:} In Table~\ref{tab:abl_GT}, we further evaluate the geometric metrics of all methods when using GT camera pose and semantic information as input. A full table with semantic evaluation is in the supplementary material. As with estimated input, \project{-S} has similar results to the corresponding fixed-size Voxblox on regions of different quality. Although the results of all methods are significantly improved, \project{-S} and \project{-SG} outperform Multi-TSDF~\cite{schmid2022panoptic} on both geometric and completion errors in the fine quality. 
Without the noise of estimated poses, objects will not be reconstructed in wrong positions. Therefore, \cite{schmid2022panoptic} cannot benefit from an incomplete reconstruction when computing the geometric error.
In the coarser region, multi-TSDFs \cite{schmid2022panoptic} achieve less completion and geometric error due to a more accurate TSDF estimation in large voxels. 
Nevertheless, this region requires less focus, since the objective is to maintain rough reconstruction on them and build a higher quality map for others.

\begin{table}[tb]
    \footnotesize
    \centering
    \resizebox{0.99\columnwidth}{!}{
    \begin{tabular}{c|cc|ccc}
        \toprule
        & \textbf{Method} & \makecell{\textbf{Reconstruction} \\ \textbf{Quality} (cm)} & \makecell{\textbf{Completion} \\ \textbf{Error} (cm) $\downarrow$} & \makecell{\textbf{Compl. $<$5cm} \\ \textbf{Ratio} (\%) $\uparrow$} & \makecell{\textbf{Geometric} \\ \textbf{Error} (cm) $\downarrow$} \\ 
        \midrule \midrule
        \parbox[t]{2mm}{\multirow{6}{*}{\rotatebox[origin=c]{90}{\textit{\red{@1cm}}}}}
        & Voxblox \cite{oleynikova2017voxblox} (fixed) & Fine [1] & \red{\textbf{\underline{0.29 $\pm$ 0.22}}} & \red{\textbf{99.99\%}} & \red{\textbf{0.36 $\pm$ 0.37}} \\
        & Multi-TSDFs \cite{schmid2022panoptic} & Multi-level [1-4-8] & \red{0.34 $\pm$ 0.56} & \red{99.71\%} & \red{0.79 $\pm$ 1.62} \\
        & \textbf{\project{-S}} & \textbf{Adaptive [1-4-8]} & \red{\textbf{0.27 $\pm$ 0.22}} & \red{\textbf{99.99\%}} & \red{0.37 $\pm$ 0.42} \\
        & \textbf{\project{-SG}} & \textbf{Adaptive [1-4-8]} & \red{0.29 $\pm$ 0.23} & \red{\textbf{99.99\%}} & \red{\textbf{\underline{0.37 $\pm$ 0.41}}} \\
        & \textit{w/o adaptive raycasting} & \textbf{Adaptive [1-4-8]} & \red{0.40 $\pm$ 0.31} & \red{99.98\%} & \red{0.38 $\pm$ 0.35} \\
        & \textit{w/o neighbor splitting} & \textbf{Adaptive [1-4-8]} & \red{0.29 $\pm$ 0.27} & \red{\textbf{\underline{99.98\%}}} & \red{0.67 $\pm$ 2.01} \\
        \midrule 
        \parbox[t]{2mm}{\multirow{6}{*}{\rotatebox[origin=c]{90}{\green{\textit{@4cm}}}}}
        & Voxblox \cite{oleynikova2017voxblox} (fixed) & Middle [4] & \green{0.92 $\pm$ 1.15} & \green{98.55\%} & \green{1.96 $\pm$ 1.96} \\
        & Multi-TSDFs \cite{schmid2022panoptic} & Multi-level [1-4-8] & \green{0.99 $\pm$ 2.39} & \green{96.85\%} & \green{1.57 $\pm$ 2.20} \\
        & \textbf{\project{-S}} & \textbf{Adaptive [1-4-8]}  & \green{0.85 $\pm$ 1.13} & \green{98.66\%} & \green{1.83 $\pm$ 1.93} \\
        &\textbf{\project{-SG}} & \textbf{Adaptive [1-4-8]} &\green{\underline{\textbf{0.49 $\pm$ 0.64}}} & \green{\underline{\textbf{99.83\%}}} & \green{\underline{\textbf{0.90 $\pm$ 1.33}}} \\
        &\textit{w/o adaptive raycasting} & \textbf{Adaptive [1-4-8]} & \green{\textbf{0.47 $\pm$ 0.54}} & \green{\textbf{99.89\%}} & \green{\textbf{0.84 $\pm$ 1.30}} \\
        &\textit{w/o neighbor splitting} & \textbf{Adaptive [1-4-8]} & \green{0.72 $\pm$ 0.88} & \green{99.59\%} & \green{1.99 $\pm$ 2.57} \\
        \midrule 
        \parbox[t]{2mm}{\multirow{6}{*}{\rotatebox[origin=c]{90}{\textit{\blue{@8cm}}}}} 
        & Voxblox \cite{oleynikova2017voxblox} (fixed) & Coarse [8] & \blue{1.43 $\pm$ 1.77} & \blue{95.54\%} & \blue{1.42 $\pm$ 2.10} \\
        & Multi-TSDFs \cite{schmid2022panoptic} & Multi-level [1-4-8] & \blue{0.87 $\pm$ 1.92} & \blue{98.13\%} & \blue{0.75 $\pm$ 1.47} \\
        & \textbf{\project{-S}} & \textbf{Adaptive [1-4-8]}  & \blue{1.32 $\pm$ 1.64} & \blue{96.33\%} & \blue{1.24 $\pm$ 1.90} \\
        & \textbf{\project{-SG}} & \textbf{Adaptive [1-4-8]}  & \blue{\textbf{0.85 $\pm$ 1.13}} & \blue{\textbf{98.90\%}} & \blue{\underline{\textbf{0.73 $\pm$ 1.18}}} \\
        & \textit{w/o adaptive raycasting} & \textbf{Adaptive [1-4-8]}  & \blue{\underline{\textbf{0.86 $\pm$ 1.14}}} & \blue{\underline{\textbf{98.87\%}}} & \blue{\textbf{0.72 $\pm$ 1.19}} \\
        & \textit{w/o neighbor splitting} & \textbf{Adaptive [1-4-8]}  & \blue{1.19 $\pm$ 1.36} & \blue{97.61\%} & \blue{1.24 $\pm$ 1.87} \\
        \bottomrule
    \end{tabular}
    }
    \caption{\textbf{Ablation Study.} Results on HSSD with GT camera pose and 2D semantic segmentation. We also investigate the impact of adaptive raycasting and neighborhood split. 
    Best values per evaluation level are in \textbf{bold}, second best in \underline{\textbf{underlined bold}}.
    }
    \label{tab:abl_GT}
    \vspace{-20pt}
\end{table}

\vspace{1mm} \noindent \textbf{Adaptive raycasting:} Table~\ref{tab:abl_GT} shows results on \project{-SG} without adaptive raycasting, \ie, using a single virtual grid for coarse level (8 cm) to decide if a point should be updated to voxels of all resolutions. Compared to using adaptive raycasting, results in coarser regions are not affected. However, completion error increases in fine regions where many holes appear. Visualization is in supplementary material.

\vspace{1mm} \noindent \textbf{Neighbor splitting:} We provide results of \project{-SG} without splitting neighboring voxels to the same resolution of a query voxel when that gets split to a finer resolution. In fine regions, although \project{-SG} without split achieves a similar completion error, it has a significantly higher geometric error due to ghost meshes generated at the boundaries of voxels in different resolutions.
\section{Conclusion}
\label{method}

We present \project{}, the first real-time quality-adaptive semantic 3D reconstruction method that creates a single map with regions of different quality levels. We showcase its performance in an end-to-end reconstruction pipeline on a simulated and a real-world dataset. When compared to baselines, it provides a lightweight semantic 3D map that is comparable or superior in geometric and semantic accuracy to using a fixed-sized map.
Compared to the only other method that creates maps of different resolutions leveraging semantic information \cite{schmid2022panoptic} -- albeit individual object-instance-based ones, our method generates more detailed and complete reconstructions without duplicate information across resolutions.

\paragraph{Acknowledgement.}
This project was supported by the ETH RobotX research grant.
{
    \small
    \bibliographystyle{ieeenat_fullname}
    \bibliography{main}
}

\clearpage
\maketitlesupplementary

\setcounter{section}{0}
\setcounter{figure}{0}
\setcounter{table}{0}
\renewcommand\thesection{\Alph{section}}
\renewcommand\thetable{\Alph{table}}
\renewcommand\thefigure{\Alph{figure}}

\begin{abstract}
In the supplementary material, we provide additional details about the following:
\begin{enumerate}
    \item Implementation details (Section \ref{sec:implementation}),
    \item Statistics on the generated scenes in HSSD (43 subscenes) and ScanNet (38 scenes) datasets (Section \ref{sec:statistics}),
    \item Additional results and ablations (Section \ref{sec:add_results}).
\end{enumerate}
\end{abstract}

\section{Implementation Details}
\label{sec:implementation}
\vspace{1mm} \noindent \textbf{Multi-resolution Mesh Generation.}
Algorithm \ref{alg:mesh_generation} summarizes how to generate meshes from our multi-resolution voxel map. For a voxel $V_p$ in the coarse resolution, we define $R(V_p)$ as the number of child voxels along each axis in $V_p$. Under the setting of three resolution levels: coarse (8 cm), middle (4 cm), and fine (1 cm), $R(V_p) \in \{1, 2, 8\}$. If $V_p$ is split , $V_{p,(i,j,k)}^{r}$ represents its child voxel of higher resolution $r \in \{middle, fine\}$, where $i,j,k \in \mathbb{N} \cap [1,R(V_p)]$ is the local index of the child voxel in $V_p$.

\begin{figure}[htbp]
    \centering
    \includegraphics[width=0.4\columnwidth, height=0.4\columnwidth]{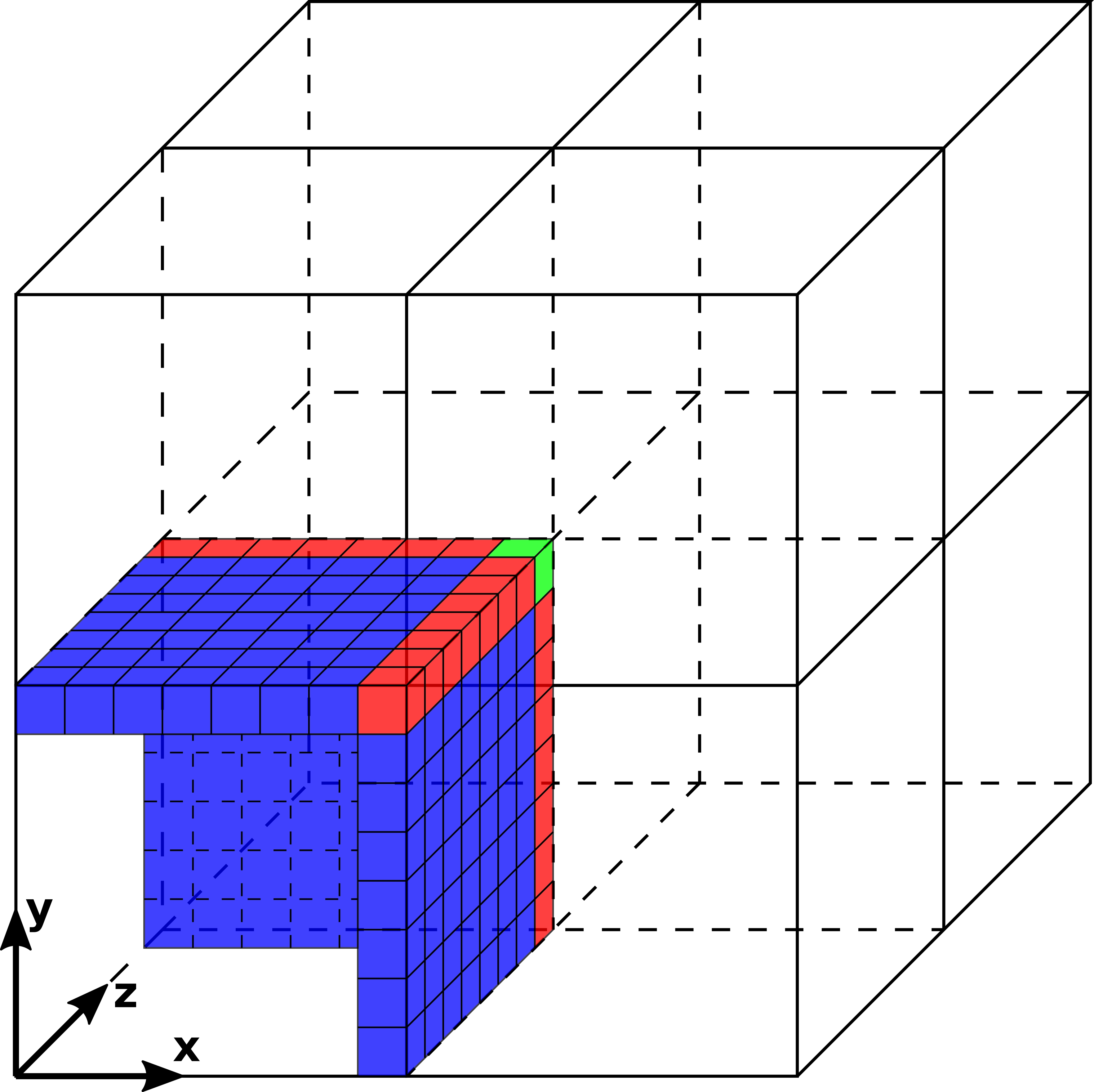}
    \caption{\textbf{Boundaries}. There are three distinct types of boundaries among coarsest voxels. \textit{(i)} To create meshes from subvoxels on \textbf{\blue{faces}}, we need adjacent subvoxels in \textbf{1} neighboring coarse voxel. \textit{(ii)} For subvoxels positioned along the \textbf{\red{edges}}, subvoxels from \textbf{3} neighbors are required. \textit{(iii)} Subvoxels from all \textbf{7} neighbors are queried to form mesh for the subvoxel located at the \textbf{\green{corner}}.}
    \label{fig:boundary}
    \vspace{-10pt}
\end{figure}

We traverse each voxel $V_p$ in the coarse resolution. When its neighbors and itself are not split, we generate meshes on these coarse-resolution voxels using \cite{lorensen1987marching}. Similar to \cite{steinbrucker2014volumetric}, there are in total 3 types of boundaries to consider between $V_p$ and its neighbors. As illustrated in Figure \ref{fig:boundary}, 3 faces, 3 edges, and 1 corner should be considered when creating mesh surfaces between them. To process these boundaries, we need to query the information from 1, 3, and 8 neighboring coarse voxels for the face, edge, and corner boundary, respectively. When $V_p$ has child voxels in resolution $r$, we traverse all of them and create mesh surfaces (line 7 in Algorithm \ref{alg:mesh_generation}), excluding those that exist on the boundaries to the neighboring coarse voxels. Last, we iterate over all these boundaries (lines 9-12 in Algorithm \ref{alg:mesh_generation}) and create meshes with the finest possible resolution among the involved coarse voxels. For child voxels that may not exist, we substitute them with coarser-resolution voxels, as discussed in the main paper.

\begin{algorithm}[htbp]
\caption{Multi-Resolution Mesh Generation}
\begin{algorithmic}[1]
\FOR{each voxel $V_{p}$ in the coarse resolution at index position $(x,y,z)\in \mathbb{N}^{3}$}
    \STATE Form a cube, $C_{p} = \{V_{p}\} \cup N_{p}$, where $ N_{p}$ is the set of $V_{p}$'s 7 neighbors at position $(x+1,y,z)$, $(x,y+1,z)$, $(x,y,z+1)$, $(x+1,y+1,z)$, $(x+1,y,z+1)$, $(x,y+1,z+1)$, $(x+1,y+1,z+1)$ $\in \mathbb{N}^{3}$.
    \IF{$\forall V_{q} \in C_{p}$,  $V_{q}$ does not have child voxels}
        \STATE Apply Marching Cubes \cite{lorensen1987marching} on $C_{p}$.
    \ELSE
        \IF{$V_{p}$ has child voxels in resolution $r$}
            \STATE Form meshes on child voxels that are inside $V_{p}$, \ie on $V_{p,(i,j,k)}^{r}$ and its 7 neighbors, where $i,j,k \in \mathbb{N} \cap [1,R(V_p)-1]$.
        \ENDIF
        \FOR{3 faces, 3 edges and 1 corner of $V_p$}
            \STATE // Assume the set of the involved coarse-size voxel is $\mathbb{B}$
            \STATE Traverse all child voxels on the boundary in resolution $\max_{V_q \in \mathbb{B}}{(R(V_q))}$ and generate meshes. If a child voxel in that resolution does not exist, substitute it with a lower-resolution voxel in the same position.
        \ENDFOR
    \ENDIF
\ENDFOR
\end{algorithmic}
\label{alg:mesh_generation}
\vspace{-2pt}
\end{algorithm}

Let us assume that we are generating meshes on $V_p$, which is at index position $(x, y, z) \in \mathbb{N}^3$, $V_q$ is its neighbor at $(x+1, y, z) \in \mathbb{N}^3$, $R(V_p) = 8$, and $R(V_q) = 2$. When processing the face boundary perpendicular to the $x$-axis, information from $V_q$ is required. The smallest possible resolution between $V_p$ and $V_q$ is $fine$ since $\max(R(V_p), R(V_q)) = 8$. Therefore, we traverse child voxels on the faces in fine resolution (\ie $V_{p,(8,j,k)}^{fine}$, where $j,k \in [1, 7]$, and generate meshes. For $V_{p,(8,j,k)}^{fine}$, its neighboring fine voxel $V_{q-(1,j,k)}^{fine}$ will be substituted by $V_{q-(1,1+j//4,1+k//4)}^{middle}$ since the smallest resolution in $V_q$ is middle.


\vspace{1mm} \noindent \textbf{Geometric Complexity.}
At each frame $k$, we do not calculate geometric complexity for all points in the projected point cloud $PC_k$, which takes more than 1s (Table \ref{tab:geo_complexity}) and prevents the whole system from running in real-time. Instead, we extract a subset of points in $PC_k$ by projecting pixels from the 2D depth map with a stride of 2. The effect of this approximation is discussed in Section \ref{sec:add_results}.
For each point $pc_j$ in the subset of $PC_k$, we form a sphere of radius $0.1$ $m$ and define other points within this sphere as the neighbors of $pc_j$. The coordinates of these neighboring points are used to form the structure tensor \cite{jutzi2009nearest} and then calculate the change of curvature \cite{pauly2002efficient,rusu2010semantic}.

\section{Dataset Statistics}
\label{sec:statistics}
\vspace{1mm} \noindent \textbf{HSSD Dataset.}
We generate one subscene for each of the 43 scenes in the validation set of the HSSD dataset \cite{khanna2023hssd}. We employ the validation set and not the test set since the scenes in the latter were not released by the time of submission. On average, each subscene covers 27.47 $m^{2}$ of area.

\begin{figure}[htb]
    \centering
    \includegraphics[width=0.99\columnwidth]{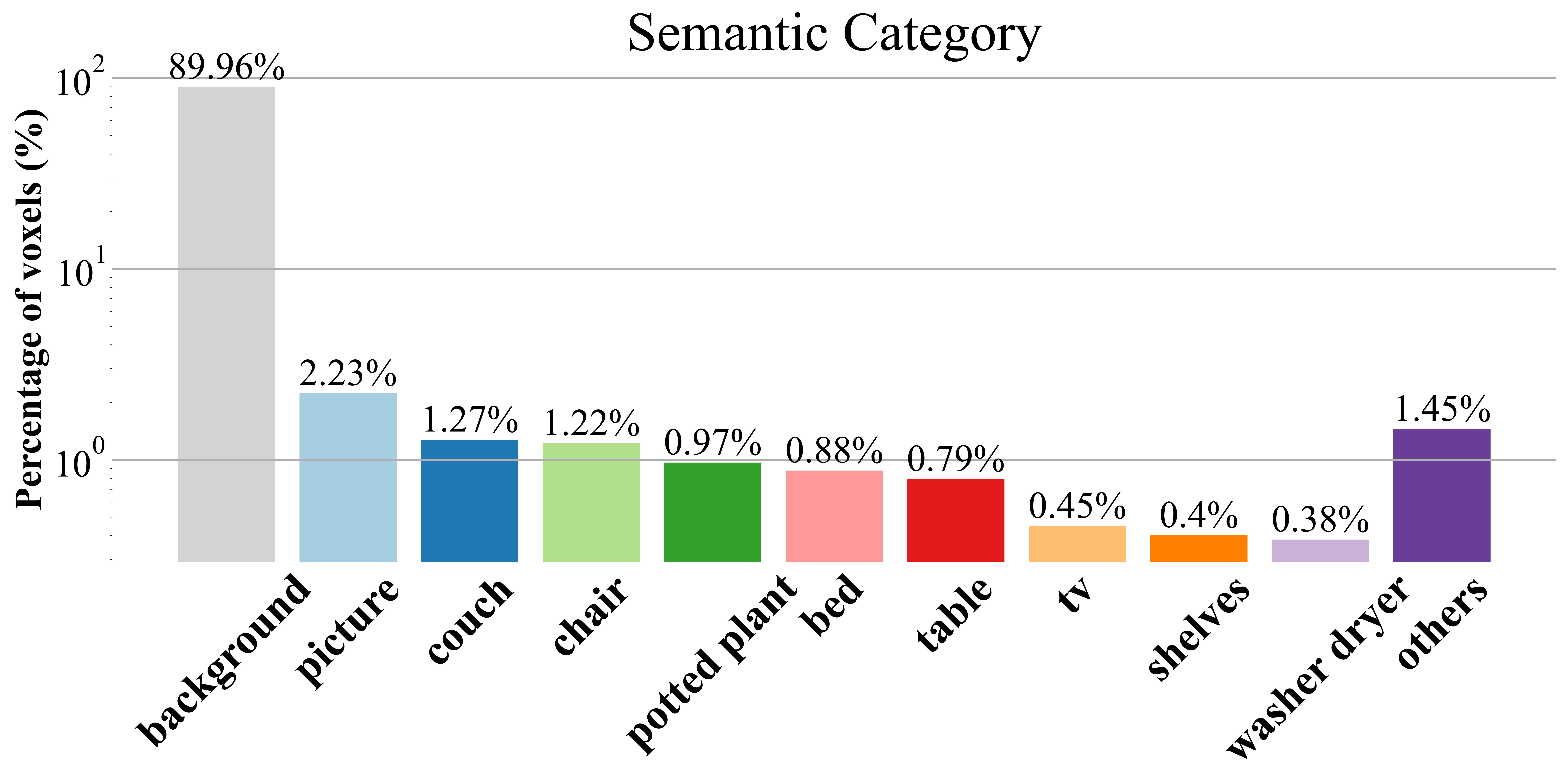}
    \caption{\textbf{Voxel-wise percentage on ground truth semantics  in the 43 subscenes from HSSD \cite{khanna2023hssd}.} Top 12 semantics are reported. Others are summarized as \textbf{\textcolor[RGB]{177,89,40}{others}}. Note that we place in `background' all map regions that do not have any semantic label in the ground truth annotations.}
    \label{fig:semantic_hssd}
    \vspace{-5pt}
\end{figure}

The sensor resolution of the RGBD frames generated by the Habitat simulation system \cite{habitat19iccv,szot2021habitat} is $640 \times 480$. The average number of frames per scene is 1550. The statistics of the semantics appearing in the subscenes are summarized in Figure \ref{fig:semantic_hssd}, where the number of voxels for all semantics is calculated on a 1cm voxel grid. Table \ref{tab:semantics_per_res_hssd} shows the semantic labels per quality level for each of the 5 random allocations we use in the experiments.

\begin{table*}[ht!]
    \footnotesize
    \centering
    \begin{tabular}{c|ccc}
    \toprule
    & \textbf{Fine [1cm]} & \textbf{Middle [4cm]} & Coarse [8cm]\\
    \midrule
    \textbf{Allocation 1} & 
    \makecell{\red{\textit{bowl, chair, chest of drawers,}} \\ \red{\textit{couch, cushion, microwave,}} \\ \red{\textit{plate, shelves, shoes, toaster}}}& 
    \makecell{\green{\textit{alarm clock, bottle, fridge, potted plant, stool,}} \\ \green{\textit{table, table lamp, tv, vase, washer dryer}}} & 
    \makecell{\blue{\textit{background, book, drinkware, laptop,}} \\ \blue{\textit{bed, picture, sink, toilet, trashcan}}}
    \\
    \midrule
    \textbf{Allocation 2}  & 
    \makecell{\red{\textit{bed, book, bottle, bowl,}} \\ \red{\textit{laptop, microwave, picture,}} \\ \red{\textit{table, vase, washer dryer}}}& 
    \makecell{\green{\textit{chest of drawers, cushion, drinkware, plate,}} \\ \green{\textit{potted plant, table lamp, toilet, trashcan, tv}}}& 
    \makecell{\blue{\textit{alarm clock, background, chair, fridge,}} \\  \blue{\textit{couch, shelves, shoes, sink, stool, toaster}}} 
    \\
    \midrule
    \textbf{Allocation 3} & 
    \makecell{\red{\textit{couch, chest of drawers, cushion,}} \\ \red{\textit{drinkware, laptop, potted plant,}} \\ \red{\textit{washer dryer, shoes, toilet, book}}}& 
    \makecell{\green{\textit{bed, bottle, chair, microwave, picture,}} \\ \green{\textit{plate, sink, table, table lamp}}} & 
    \makecell{\blue{\textit{alarm clock, background, bowl, fridge,}} \\  \blue{\textit{shelves, stool, toaster, trashcan, tv, vase}}} 
    \\
    \midrule
    \textbf{Allocation 4} & 
    \makecell{\red{\textit{alarm clock, book, chair,}} \\ \red{\textit{couch, cushion, drinkware, shoes,}} \\ \red{\textit{sink, table lamp, washer dryer}}} & 
    \makecell{\green{\textit{bottle, chest of drawers, fridge, microwave,}} \\ \green{\textit{potted plant, shelves, stool, toilet, trashcan}}} & 
    \makecell{\blue{\textit{background, bed, bowl, laptop, picture,}} \\  \blue{\textit{plate, table, toaster, tv, vase}}} 
    \\
    \midrule
    \textbf{Allocation 5} & 
    \makecell{\red{\textit{alarm clock, chair, fridge,}} \\ \red{\textit{laptop, picture, plate, shoes,}} \\ \red{\textit{stool, toilet, washer dryer}}} & 
    \makecell{\green{\textit{bowl, couch, microwave, potted plant,}} \\ \green{\textit{sink, table, trashcan, tv, vase}}} & 
    \makecell{\blue{\textit{background, bed, book, bottle, chest of drawers,}} \\  \blue{\textit{cushion, drinkware, shelves, table lamp, toaster}}} 
    \\
    \bottomrule
    \end{tabular}
    \caption{\textbf{Allocation of semantic classes per resolution category on HSSD \cite{khanna2023hssd}.}}
    \label{tab:semantics_per_res_hssd}
    \vspace{-5pt}
\end{table*}

\begin{figure}[htb]
    \centering
    \includegraphics[width=0.99\columnwidth]{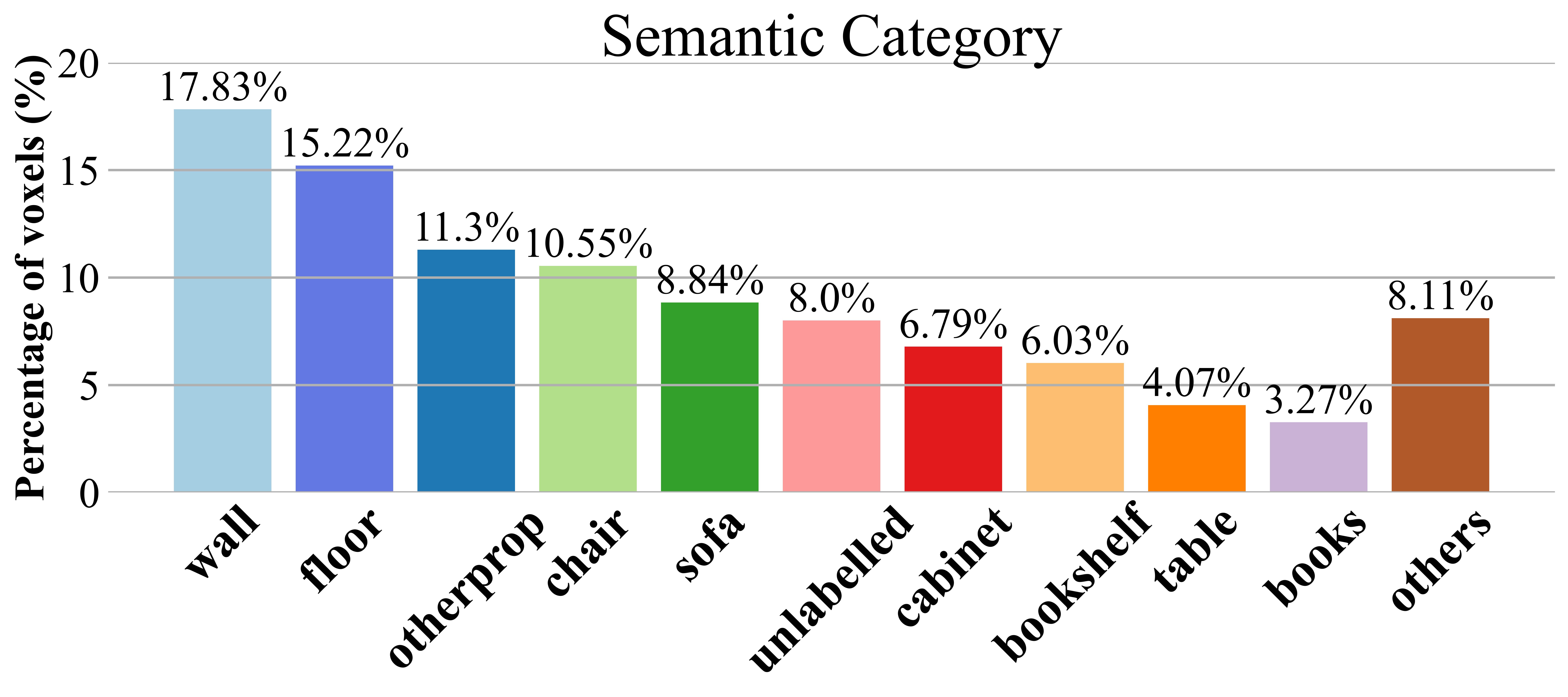}
    \caption{\textbf{Voxel-wise percentage on ground truth semantics  in the 38 scenes from ScanNet \cite{dai2017scannet}.} Top 10 semantics are reported. Others are summarized as \textbf{\textcolor[RGB]{177,89,40}{others}}.}
    \label{fig:semantic_scannet}
    \vspace{-5pt}
\end{figure}

\vspace{1mm} \noindent \textbf{ScanNet Dataset.}
We randomly select 38 scans of scenes from the validation set\footnote{We employ the validation set since the test set does not have publicly available annotations.} of ScanNet. 
We provide the list of the selected scenes (\textit{scannet\_scenes/scannetv2\_val\_picked.txt}) and the code to generate it (\textit{scannet\_scenes/random\_pick.py}). 
We initially sample 40 scenes.
However, \cite{schmid2022panoptic} fails in two of the scans (\textit{scene0678\_01} and \textit{scene0231\_01}) due to out of memory.
Therefore, we eliminate these 2 scenes, resulting in a total of 38 scenes.
The resolution of the depth images in the ScanNet dataset is $640 \times 480$.
We downsample the RGB images, whose original resolution is $1296 \times 968$, to be the same size as the depth images by bicubic interpolation.
The statistics of semantics over all 38 scenes are shown in Figure \ref{fig:semantic_scannet}.
The allocation of semantic categories per quality level can be found in Table \ref{tab:semantics_per_res_scannet}.
Please note that we followed, to the extent possible, a similar allocation to the randomly selected labels in HSSD, to provide more comparable results between the two datasets.

\begin{table*}[ht!]
    \footnotesize
    \centering
    \begin{tabular}{c|ccc}
    \toprule
    & \textbf{Fine [1cm]} & \textbf{Middle [4cm]} & \textbf{Coarse [8cm]}\\
    \midrule
    \textbf{Allocation 1} & 
    \makecell{\red{\textit{cabinet, night stand, bookshelf,}} \\ \red{\textit{sofa, desk, shelves, dresser,}} \\ \red{\textit{pillow, mirror, clothes, towel,}} \\ \red{\textit{person, door, other furniture}}}& 
    \makecell{\green{\textit{chair, table, counter, blinds, curtain,}} \\ \green{\textit{refridgerator, television, shower curtain,}} \\ \green{\textit{box, lamp, bathtub, paper, bag, otherprop}}} & 
    \makecell{\blue{\textit{wall, floor, bed, window,}} \\ \blue{\textit{picture, floor mat, ceiling, books,}} \\ \blue{\textit{whiteboard, toilet, sink, other structure}}} 
    \\
    \midrule
    \textbf{Allocation 2}  & 
    \makecell{\red{\textit{cabinet, bed, bag, table,}} \\ \red{\textit{window, picture, blinds, curtain,}} \\ \red{\textit{mirror, floor mat, books, box,}} \\ \red{\textit{whiteboard, bathtub, other furniture}}}& 
    \makecell{\green{\textit{counter, towel, desk, dresser, pillow,}} \\ \green{\textit{television, shower curtain, night stand,}} \\ \green{\textit{toilet, lamp, other structure, otherprop}}} & 
    \makecell{\blue{\textit{wall, floor, chair, sofa,}} \\ \blue{\textit{door, bookshelf, shelves, clothes,}} \\ \blue{\textit{ceiling, refridgerator, paper, person, sink}}} 
    \\
    \midrule
    \textbf{Allocation 3} & 
    \makecell{\red{\textit{cabinet, sofa, bookshelf,}} \\ \red{\textit{dresser, pillow, mirror,}} \\ \red{\textit{clothes, books, whiteboard,}} \\ \red{\textit{person, night stand, toilet}}}& 
    \makecell{\green{\textit{bed, towel, table, window, picture,}} \\ \green{\textit{counter, curtain, lamp, sink, paper,}} \\ \green{\textit{shower curtain, floor mat, other structure}}} & 
    \makecell{\blue{\textit{wall, floor, chair, box, door, refridgerator,}} \\ \blue{\textit{desk, shelves, ceiling, blinds, television,}} \\ \blue{\textit{bathtub, bag, other furniture, otherprop}}} 
    \\
    \midrule
    \textbf{Allocation 4} & 
    \makecell{\red{\textit{cabinet, sofa, counter,}} \\ \red{\textit{desk, bag, pillow, sink,}} \\ \red{\textit{mirror, clothes, books, lamp,}} \\ \red{\textit{person, curtain, other furniture}}}& 
    \makecell{\green{\textit{chair, window, bookshelf, blinds, dresser,}} \\ \green{\textit{shelves, whiteboard, floor mat, toilet, paper,}} \\ \green{\textit{box, refridgerator, other structure, otherprop}}} & 
    \makecell{\blue{\textit{wall, floor, bed, table, door,}} \\ \blue{\textit{picture, night stand, television,}} \\ \blue{\textit{towel, shower curtain, ceiling, bathtub}}} 
    \\
    \midrule
    \textbf{Allocation 5} & 
    \makecell{\red{\textit{cabinet, chair, bookshelf,}} \\ \red{\textit{picture, counter, desk,}} \\ \red{\textit{refridgerator, night stand,}} \\ \red{\textit{clothes, toilet, otherprop}}}& 
    \makecell{\green{\textit{sofa, table, door, blinds, box,}} \\ \green{\textit{television, paper, shower curtain, whiteboard,}} \\ \green{\textit{person, sink, other structure, other furniture}}} & 
    \makecell{\blue{\textit{wall, floor, bed, window, bag, shelves,}} \\ \blue{\textit{curtain, dresser, pillow, lamp, mirror,}} \\ \blue{\textit{floor mat, ceiling, books, towel, bathtub}}} 
    \\
    \bottomrule
    \end{tabular}
    \caption{\textbf{Allocation of semantic classes per resolution category on ScanNet \cite{dai2017scannet}.}}
    \label{tab:semantics_per_res_scannet}
    \vspace{-5pt}
\end{table*}

\section{Additional Results}
\label{sec:add_results}
\vspace{1mm} \noindent \textbf{HSSD Dataset.}
In Table \ref{tab:hssd_full}, we show the full geometric and semantic evaluation on all 43 subscenes generated from HSSD. Results of the fixed-size method (Voxblox) \cite{oleynikova2017voxblox} on regions that do not correspond to its voxel size, \eg results of the Voxblox (1 cm) at \textit{\blue{Eval @8cm}}, are reported in \grey{grey}. For comparison, we show the performance of multi-resolution methods as well. This means that all \red{red}, \green{green}, and \blue{blue} values are from the main paper and all \grey{grey} are newly reported here. As expected, reconstructing everything at the finest level, \ie, Voxblox (1 cm) performs the best in terms of completion error in all evaluation regions. However, it has the greatest geometric error at \textit{\blue{Eval @8cm}}. As explained in the main paper, this is due to the inaccurate estimated pose, which causes reconstructed objects to appear in wrong positions. We can also observe that the geometric error decreases when the resolution quality of the fixed-size Voxblox decreases. A similar behavior can be seen for semantic evaluation, where the coarsest Voxblox provides the best results, a performance that degrades when the resolution quality increases.

\begin{table*}[ht!]
    \footnotesize
    \centering
    \resizebox{\textwidth}{!}{
    \begin{tabular}{c|cc|ccccc}
        \toprule
          & \textbf{Method} & \makecell{\textbf{Reconstruction} \\ \textbf{Quality} [cm]} & \makecell{\textbf{Completion} \\ \textbf{Error} (cm) $\downarrow$} & \makecell{\textbf{Completion $<$5cm} \\ \textbf{Ratio} (\%) $\uparrow$} & \makecell{\textbf{Geometric} \\ \textbf{Error} (cm) $\downarrow$} & \makecell{\textbf{Semantic} \\ \textbf{Accuracy} (\%) $\uparrow$} & \makecell{\textbf{Semantic} \\ \textbf{mIoU} (\%) $\uparrow$} \\ 
         \midrule \midrule

         \parbox[t]{2mm}{\multirow{6}{*}{\rotatebox[origin=c]{90}{\textit{\red{Eval @1cm}}}}}
         & \red{Voxblox} \cite{oleynikova2017voxblox} & \red{Fine [1]} & \textbf{\red{2.49 $\pm$ 2.80}} & \textbf{\red{88.74}} & \red{4.14 $\pm$ 4.49} & \red{12.96} & \phantom{1}\red{6.62}\\
         & \grey{Voxblox} \cite{oleynikova2017voxblox} & \grey{Middle [4]} & \grey{3.05 $\pm$ 3.16} & \grey{83.49} & \grey{4.16 $\pm$ 4.50} & \grey{37.93} & \grey{16.05}\\
         & \grey{Voxblox} \cite{oleynikova2017voxblox} & \grey{Coarse [8]} &  \grey{3.67 $\pm$ 3.57} & \grey{74.77} & \grey{5.28 $\pm$ 5.11} & \textbf{\grey{51.78}} & \textbf{ \grey{22.18}} \\ 
         & \red{Multi-TSDFs} \cite{schmid2022panoptic} & \red{Multi-level [1-4-8]} & \red{2.74 $\pm$ 4.00} & \red{85.59} & \textbf{\red{4.10 $\pm$ 6.53}} & \phantom{1}\red{8.58} & \phantom{1}\red{4.86} \\
         & \red{\textbf{MAP-ADAPT-S}} & \red{Adaptive [1-4-8]} & \red{2.54 $\pm$ 2.92} & \red{88.15} & \red{4.18 $\pm$ 4.62} & \red{13.12}  & \phantom{1}\red{6.74}\\
         & \red{\textbf{MAP-ADAPT-SG}} & \red{Adaptive [1-4-8]} & \red{2.53 $\pm$ 2.84} & \red{88.34} & \red{4.19 $\pm$ 4.57} & \red{13.12}  & \phantom{1}\red{6.74}\\
        \midrule
        \parbox[t]{2mm}{\multirow{6}{*}{\rotatebox[origin=c]{90}{\green{\textit{Eval @4cm}}}}}
         & \grey{Voxblox} \cite{oleynikova2017voxblox} & \grey{Fine [1]} & \textbf{\grey{2.44 $\pm$ 2.97}} & \textbf{\grey{89.91}} & \grey{3.91 $\pm$ 4.19} & \grey{12.45} & \phantom{1}\grey{6.88}\\
         & \green{Voxblox} \cite{oleynikova2017voxblox} & \green{Middle [4]} & \green{3.06 $\pm$ 3.50} & \green{84.39} & \green{4.10 $\pm$ 4.16} & \green{40.00} & \green{16.01}\\
         & \grey{Voxblox} \cite{oleynikova2017voxblox} & \grey{Coarse [8]} & \grey{3.80 $\pm$ 4.02} & \grey{74.51} & \grey{5.28 $\pm$ 4.86} & \textbf{\grey{55.49}} & \textbf{\grey{21.40}} \\
         & \green{Multi-TSDFs} \cite{schmid2022panoptic} & \green{Multi-level [1-4-8]} & \green{3.09 $\pm$ 3.83} & \green{83.29} & \green{4.18 $\pm$ 6.46} & \green{10.57} & \phantom{1}\green{6.41} \\
         & \green{\textbf{MAP-ADAPT-S}} & \green{Adaptive [1-4-8]} & \green{2.89 $\pm$ 3.45} & \green{86.12} & \green{4.05 $\pm$ 4.19} & \green{39.69}  & \green{16.26}\\
         & \green{\textbf{MAP-ADAPT-SG}} & \green{Adaptive [1-4-8]} & \green{2.67 $\pm$ 3.25} & \green{88.04} & \textbf{\green{3.85 $\pm$ 4.13}} & \green{39.88}  & \green{16.21}\\
        \midrule
        \parbox[t]{2mm}{\multirow{6}{*}{\rotatebox[origin=c]{90}{\textit{\blue{Eval @8cm}}}}} 
         & \grey{Voxblox} \cite{oleynikova2017voxblox} & \grey{Fine [1]} & \textbf{\grey{2.39 $\pm$ 2.71}} & \textbf{\grey{89.79}} & \grey{5.11 $\pm$ 6.37} & \grey{16.23} & \phantom{1}\grey{6.25}\\
         & \grey{Voxblox} \cite{oleynikova2017voxblox} & \grey{Middle [4]} & \grey{3.03 $\pm$ 3.13} & \grey{83.97} & \grey{4.64 $\pm$ 6.35} & \grey{44.19} & \grey{15.66} \\
         & \blue{Voxblox} \cite{oleynikova2017voxblox} & \blue{Coarse [8]} & \blue{3.59 $\pm$ 3.59} & \blue{77.93} & \blue{4.57 $\pm$ 6.11} & \textbf{\blue{60.38}} & \textbf{\blue{21.46}} \\ 
         & \blue{Multi-TSDFs} \cite{schmid2022panoptic} & \blue{Multi-level [1-4-8]} & \blue{3.42 $\pm$ 3.79} & \blue{79.86} & \textbf{\blue{4.05 $\pm$ 5.89}} & \blue{49.59} & \phantom{1}\blue{8.85} \\
         & \blue{\textbf{MAP-ADAPT-S}} & \blue{Adaptive [1-4-8]} & \blue{3.43 $\pm$ 3.47} & \blue{79.94} & \blue{4.53 $\pm$ 5.95} & \textbf{\blue{60.38}}  & \blue{21.18}\\
         & \blue{\textbf{MAP-ADAPT-SG}} & \blue{Adaptive [1-4-8]} & \blue{3.10 $\pm$ 3.27} & \blue{83.56} & \blue{4.53 $\pm$ 5.89} & \textbf{\blue{60.38}}  & \blue{21.17}\\
        \bottomrule
    \end{tabular}
    }
    \caption{\textbf{Evaluation per quality level on \textbf{\textit{all 43 scenes}} from HSSD \cite{khanna2023hssd}.} Best values per evaluation level are in \textbf{bold}. We consider only directly-comparable methods per resolution and report everything else \grey{in grey} for reference purposes.}
    \label{tab:hssd_full}
\end{table*}

\vspace{1mm} \noindent \textbf{ScanNet Dataset.}
The corresponding full results on the ScanNet dataset \cite{dai2017scannet} of Voxblox \cite{oleynikova2017voxblox} on all resolutions, Multi-TSDF \cite{schmid2022panoptic}, and our method are reported in Table \ref{tab:scannet_full}, from which similar observations can be made to the results on HSSD. We also report the average running time and memory size per method in Table \ref{tab:runtime_scannet}. \project{} still occupies less memory than Voxblox (1cm) and runs significantly faster than \cite{schmid2022panoptic} by a clear margin. Since the proportion of regions belonging to coarse quality in ScanNet is less than in HSSD, the amount of saved map size is not as substantial as in HSSD. We provide further analysis with the percentage of each quality level in the next section.

\begin{table*}[ht!]
    \footnotesize
    \centering
    \resizebox{\textwidth}{!}{
    \begin{tabular}{c|cc|ccccc}
        \toprule
          & \textbf{Method} & \makecell{\textbf{Reconstruction} \\ \textbf{Quality} [cm]} & \makecell{\textbf{Completion} \\ \textbf{Error} (cm) $\downarrow$} & \makecell{\textbf{Completion $<$5cm} \\ \textbf{Ratio} (\%) $\uparrow$} & \makecell{\textbf{Geometric} \\ \textbf{Error} (cm) $\downarrow$} & \makecell{\textbf{Semantic} \\ \textbf{Accuracy} (\%) $\uparrow$} & \makecell{\textbf{Semantic} \\ \textbf{mIoU} (\%) $\uparrow$} \\ 
         \midrule \midrule

         \parbox[t]{2mm}{\multirow{6}{*}{\rotatebox[origin=c]{90}{\textit{\red{Eval @1cm}}}}}
         & \red{Voxblox} \cite{oleynikova2017voxblox} & \red{Fine [1]} & \textbf{\red{3.21 $\pm$ 4.92}} & \textbf{\red{82.61}} & \phantom{1}\red{7.08 $\pm$ 13.38} & \red{10.36} & \red{6.60} \\
         & \grey{Voxblox} \cite{oleynikova2017voxblox} & \grey{Middle [4]} & \grey{4.54 $\pm$ 5.66} & \grey{69.97} & \grey{\phantom{1}6.41 $\pm$ 11.97} & \grey{10.14} & \grey{5.67}\\
         & \grey{Voxblox} \cite{oleynikova2017voxblox} & \grey{Coarse [8]} &  \grey{5.30 $\pm$ 6.10} & \grey{63.46} & \grey{\phantom{1}6.75 $\pm$ 11.74} & \textbf{\grey{13.91}} & \textbf{\grey{7.34}} \\ 
         & \red{Multi-TSDFs} \cite{schmid2022panoptic} & \red{Multi-level [1-4-8]} & \red{3.75 $\pm$ 5.69} & \red{77.42} & \phantom{1}\textbf{\red{5.53 $\pm$ 10.47}} & \phantom{1}\red{6.55} & \red{4.41} \\
         & \red{\textbf{MAP-ADAPT-S}} & \red{Adaptive [1-4-8]} & \red{3.36 $\pm$ 5.20} & \red{81.57} & \phantom{1}\red{6.31 $\pm$ 11.51} & \red{10.40} & \red{6.60} \\
         & \red{\textbf{MAP-ADAPT-SG}} & \red{Adaptive [1-4-8]} & \red{3.27 $\pm$ 5.03} & \red{82.27} & \phantom{1}\red{6.84 $\pm$ 12.99} & \red{10.36} & \red{6.59} \\
        \midrule
        \parbox[t]{2mm}{\multirow{6}{*}{\rotatebox[origin=c]{90}{\green{\textit{Eval @4cm}}}}}
         & \grey{Voxblox} \cite{oleynikova2017voxblox} & \grey{Fine [1]} & \textbf{\grey{3.63 $\pm$ 5.81}} & \textbf{\grey{80.89}} & \grey{\phantom{1}8.88 $\pm$ 16.30} & \phantom{1}\grey{7.90} & \grey{6.04}\\
         & \green{Voxblox} \cite{oleynikova2017voxblox} & \green{Middle [4]} & \green{4.93 $\pm$ 6.52} & \green{69.52} & \phantom{1}\green{7.90 $\pm$ 14.80} & \phantom{1}\green{9.07} & \green{5.76} \\
         & \grey{Voxblox} \cite{oleynikova2017voxblox} & \grey{Coarse [8]} & \grey{5.74 $\pm$ 6.86} & \grey{61.93} & \grey{\phantom{1}8.18 $\pm$ 14.40} & \textbf{\grey{12.53}} & \textbf{\grey{7.12}} \\
         & \green{Multi-TSDFs} \cite{schmid2022panoptic} & \green{Multi-level [1-4-8]} & \green{4.43 $\pm$ 6.80} & \green{74.94} & \phantom{1}\textbf{\green{6.95 $\pm$ 13.28}} & \phantom{1}\green{4.47} & \green{3.23} \\
         & \green{\textbf{MAP-ADAPT-S}} & \green{Adaptive [1-4-8]}  & \green{4.24 $\pm$ 6.22} & \green{75.62} & \phantom{1}\green{8.02 $\pm$ 14.84} & \phantom{1}\green{8.89} & \green{5.71} \\
         & \green{\textbf{MAP-ADAPT-SG}} & \green{Adaptive [1-4-8]} & \green{3.91 $\pm$ 6.02} & \green{78.82} & \phantom{1}\green{8.58 $\pm$ 16.20} & \phantom{1}\green{9.00} & \green{5.73} \\
        \midrule
        \parbox[t]{2mm}{\multirow{6}{*}{\rotatebox[origin=c]{90}{\textit{\blue{Eval @8cm}}}}} 
         & \grey{Voxblox} \cite{oleynikova2017voxblox} & \grey{Fine [1]} & \textbf{\grey{3.79 $\pm$ 5.94}} & \textbf{\grey{78.95}} & \grey{\phantom{1}9.47 $\pm$ 15.50} & \grey{12.72} & \grey{7.07}\\
         & \grey{Voxblox} \cite{oleynikova2017voxblox} & \grey{Middle [4]} & \grey{5.51 $\pm$ 6.68} & \grey{62.58} & \grey{11.92 $\pm$ 18.32} & \grey{13.89} & \grey{6.75} \\
         & \blue{Voxblox} \cite{oleynikova2017voxblox} & \blue{Coarse [8]} & \blue{6.48 $\pm$ 7.30} & \blue{55.36} & \blue{11.43 $\pm$ 17.92} & \textbf{\blue{19.05}} & \textbf{\blue{8.94}} \\ 
         & \blue{Multi-TSDFs} \cite{schmid2022panoptic} & \blue{Multi-level [1-4-8]} & \blue{5.23 $\pm$ 7.02} & \blue{67.00} & \phantom{1}\textbf{\blue{9.02 $\pm$ 15.02}} & \blue{14.10} & \blue{5.28} \\
         & \blue{\textbf{MAP-ADAPT-S}} & \blue{Adaptive [1-4-8]} & \blue{5.01 $\pm$ 6.64} & \blue{67.77} & \phantom{1}\blue{9.94 $\pm$ 16.07} & \textbf{\blue{19.05}} & \textbf{\blue{8.94}} \\
         & \blue{\textbf{MAP-ADAPT-SG}} & \blue{Adaptive [1-4-8]} & \blue{4.27 $\pm$ 6.19} & \blue{74.51} & \phantom{1}\blue{9.48 $\pm$ 15.82} & \textbf{\blue{19.05}} & \textbf{\blue{8.94}} \\
        \bottomrule
    \end{tabular}
    }
    \caption{\textbf{Evaluation per quality level on \textit{all 38 scenes} from ScanNet \cite{dai2017scannet}.} Best values per evaluation level are in \textbf{bold}. We consider only directly-comparable methods per resolution and report everything else \grey{in grey} for reference purposes.}
    \label{tab:scannet_full}
\end{table*}

\begin{table}[htbp!]
    \centering
    \resizebox{0.99\columnwidth}{!}{
    \begin{tabular}{cc|ccc}
        \toprule
        \multirow{2}{*}{\textbf{Method}} & \multirow{2}{*}{\makecell{\textbf{Map Size} \\ (MB) $\downarrow$}} & \multicolumn{2}{c}{\textbf{Runtime} (ms) $\downarrow$} \\
         & & Update TSDF & Generate Mesh \\
        \midrule \midrule
        Voxblox~\cite{oleynikova2017voxblox} (1cm) & 905.15 & \phantom{1}55.37 $\pm$ \phantom{1}\phantom{1}7.46 & \phantom{1}499.13 $\pm$ \phantom{1}433.16 \\
        Voxblox~\cite{oleynikova2017voxblox} (4cm) & \phantom{1}42.24 & \phantom{1}36.20 $\pm$ \phantom{1}\phantom{1}3.14 & \phantom{1}\phantom{1}21.24 $\pm$ \phantom{1}\phantom{1}13.34 \\
        Voxblox~\cite{oleynikova2017voxblox} (8cm) & \phantom{1}\phantom{1}\textbf{10.67} & \phantom{1}\textbf{34.45 $\pm$ \phantom{1}\phantom{1}2.83} & \textbf{\phantom{1}\phantom{1}\phantom{1}6.12 $\pm$ \phantom{1}\phantom{1}\phantom{1}3.18}\\
        \midrule
        Multi-TSDFs \cite{schmid2022panoptic} & \underline{\textbf{415.38}} & 174.44 $\pm$ 132.80 & \phantom{1}\underline{\textbf{345.08 $\pm$ \phantom{1}734.73}} \\        
        \textbf{\project{-S}} & 472.02 & \phantom{1}\underline{\textbf{49.41 $\pm$ \phantom{1}\phantom{1}5.88}} & 1277.41 $\pm$ \phantom{1}896.99 \\
        \textbf{\project{-SG}} & 746.07 & \phantom{1}53.96 $\pm$ \phantom{1}\phantom{1}7.76 & 2239.91 $\pm$ 1692.50 \\
        \bottomrule
    \end{tabular}
    }
    \vspace{-6pt}
    \caption{\textbf{Map size and runtime ScanNet \cite{dai2017scannet}.} \textit{Best} values are \textbf{bold}. \textit{Best} of multi-resolution methods are in \underline{\textbf{underlined bold}}. Note that update TSDF is processed at each frame, whereas mesh generation only needs to be executed once at the end.}
    \label{tab:runtime_scannet}
    \vspace{-12pt}
\end{table}

\vspace{1mm} \noindent \textbf{Volumetric Percentage per Quality Level.}
The volumetric percentage of each quality level on both datasets is reported in Table \ref{tab:voxel_percentage}. The volumetric percentage is calculated by scaling the number of voxels in each quality level by their corresponding size (e.g., $1$ \blue{coarse (8cm)} voxel takes the same volume as $2 \times 2 \times 2$ \green{middle (4cm)} voxels). For HSSD, $75.2\%$ of the volume of the scenes is reconstructed in the coarse level by \project{-S} since most regions in this dataset are classified as background, which belongs to the coarse level in all 5 splits. Due to the noisy 2D semantic segmentation and the design of splitting neighbors, this percentage value is lower than the ground truth ratio of regions classified as background ($89.96 \%$ in Figure \ref{fig:semantic_hssd}). More areas are reconstructed in middle and fine level by \project{-SG} since this method leverages geometric complexity as an additional criterion to split voxels.

In the ScanNet dataset, nearly half of the scenes' volume is reconstructed in fine quality level by \project{-S}. This explains the decrease for \project{-S} on runtime for updating TSDF values and map size compared to Voxblox (1cm), as reported in Table \ref{tab:runtime_scannet}. Moreover, the depth frames in this dataset are measured from real-world sensors, hence the depth values are noisy and, thus, cause an overestimation of geometric complexity. As a result, $28.5\%$ more regions are reconstructed in fine quality by \project{-SG}, which uses geometric complexity as an additional voxel-split criterion. Since $70\%$ of space is reconstructed in the fine level (1cm) by \project{-SG}, it only takes slightly less map size and spends nearly the same time updating TSDF voxels compared to Voxblox (1cm). Note that \project{-SG} requires additional running time to split / merge voxels and integrate the geometric complexity of the new frames.

\begin{table}[ht!]
    \footnotesize
    \centering
    \begin{tabular}{c|c|ccc}
        \toprule
        &\multirow{2}{*}{\textbf{Method}} & \multicolumn{3}{c}{\textbf{Volume percentage of voxels}} \\
         & & \red{Fine} & \green{Middle} & \blue{Coarse} \\
        \midrule
        \multirow{2}{*}{\textbf{HSSD} \cite{khanna2023hssd}}
        & \project{-S} & 14.1\% & 10.7\% & 75.2\% \\
        & \project{-SG} & 26.9\% & 20.5\% & 52.6\% \\
        \midrule
        \multirow{2}{*}{\textbf{ScanNet} \cite{dai2017scannet}}
        & \project{-S} & 41.5\% & 21.5\% & 37.0\% \\ 
        & \project{-SG} & 70.0\% & 15.4\% & 14.6\% \\
        \bottomrule
    \end{tabular}
    \caption{\textbf{Volumetric percentage of voxels per quality level.} The volumetric percentage is calculated by scaling the number of voxels in different quality levels by their corresponding size -- e.g., $1$ \blue{coarse (8cm)} voxel takes the same volume as $2 \times 2 \times 2$ \green{middle (4cm)} voxels.}
    \label{tab:voxel_percentage}
\end{table}

\vspace{1mm} \noindent \textbf{Allocating Quality Level of Semantics Based on Physical Size.}
We conduct an experiment on the HSSD dataset \cite{khanna2023hssd} with a new quality split where the quality level of each category is determined by its physical size. We use the diagonal length of the bounding box reported in Figure 11 of \cite{khanna2023hssd} as the criterion for deciding the quality level. Semantics whose average diagonal length is less than 1m are categorized as fine level, those that are longer than 2m are classified as coarse level, and all the remaining semantics are in middle.

The evaluation results of the new quality split are reported in Table \ref{tab:result_new_allocation}. Consistent with the results of the random quality allocations reported in the main paper, our method has better or comparable reconstructions to fixed-size Voxblox. In addition, our method outperforms Multi-TSDFs \cite{schmid2022panoptic} in terms of completion error and semantic accuracy, especially in regions where semantics are at the fine level (\textit{Eval @ 1cm}). The results showcase the robustness of our method in different quality-level splits. 

\begin{table}[ht!]
    \vspace{-6pt}
    \centering
    \resizebox{\columnwidth}{!}{
    \begin{tabular}{c|c|ccccc}
        \toprule
        &\textbf{Method} & \makecell{\textbf{Compl.} \\ \textbf{Error} (cm) $\downarrow$} & \makecell{\textbf{Compl. $<$5cm} \\ \textbf{Ratio} (\%) $\uparrow$} & \makecell{\textbf{Geom.} \\ \textbf{Error} (cm) $\downarrow$} & \makecell{\textbf{Sem.} \\ \textbf{Acc.} (\%) $\uparrow$} & \makecell{\textbf{Sem.} \\ \textbf{mIoU} (\%) $\uparrow$} \\ 
        \midrule 
        \parbox[t]{2mm}{\multirow{4}{*}{\rotatebox[origin=c]{90}{\textit{\red{Eval @1cm}}}}} 
        &Voxblox [1cm]  & \textbf{\red{2.13 $\pm$ 2.53}} & \textbf{\red{93.19}} & \textbf{\red{3.43 $\pm$ 3.33}} & \red{12.88} & \phantom{1}\red{5.11} \\
        &Multi-TSDFs & \red{3.05 $\pm$ 4.07} & \red{83.97} & \red{3.88 $\pm$ 6.34} & \phantom{1}\red{5.64} & \phantom{1}\red{3.19} \\
        &\textbf{\project{-S}} & \red{2.27 $\pm$ 3.49} & \red{92.21} & \red{3.45 $\pm$ 3.27} & \textbf{\red{13.10}} & \phantom{1}\red{5.28} \\
        &\textbf{\project{-SG}} & \red{2.17 $\pm$ 2.57} & \red{92.67} & \red{3.44 $\pm$ 3.23} & \red{13.05} & \textbf{\phantom{1}\red{5.30}} \\
        \midrule 
        \parbox[t]{2mm}{\multirow{4}{*}{\rotatebox[origin=c]{90}{\textit{\green{Eval @4cm}}}}} 
        &Voxblox [4cm] & \green{2.91 $\pm$ 3.40} & \green{86.55} & \green{4.55 $\pm$ 5.21} & \textbf{\green{35.22}} & \green{19.11} \\
        &Multi-TSDFs & \green{3.09 $\pm$ 3.86} & \green{83.71} & \green{4.48 $\pm$ 6.38} & \green{10.05} & \phantom{1}\green{8.13} \\
        &\textbf{\project{-S}}  & \green{2.86 $\pm$ 3.40} & \green{86.79} & \green{4.58 $\pm$ 5.28} & \green{34.97} & \green{19.14} \\
        &\textbf{\project{-SG}} & \textbf{\green{2.76 $\pm$ 3.35}} & \textbf{\green{87.91}} & \textbf{\green{4.44 $\pm$ 5.15}} & \green{35.14} & \textbf{\green{19.16}} \\
        \midrule 
        \parbox[t]{2mm}{\multirow{4}{*}{\rotatebox[origin=c]{90}{\textit{\blue{Eval @8cm}}}}} 
        &Voxblox [8cm]  & \blue{3.61 $\pm$ 3.59} & \blue{77.65} & \blue{4.54 $\pm$ 6.07} & \textbf{\blue{60.41}} & \blue{33.68} \\
        &Multi-TSDFs & \blue{3.43 $\pm$ 3.77} & \blue{79.67} & \textbf{\blue{4.02 $\pm$ 5.94}} & \blue{49.20} & \blue{17.07} \\
        &\textbf{\project{-S}} & \blue{3.45 $\pm$ 3.48} & \blue{79.58} & \blue{4.41 $\pm$ 5.74} & \textbf{\blue{60.41}} & \textbf{\blue{33.90}} \\
        &\textbf{\project{-SG}} & \textbf{\blue{3.12 $\pm$ 3.27}} & \textbf{\blue{83.26}} & \blue{4.42 $\pm$ 5.67} & \textbf{\blue{60.41}} & \textbf{\blue{33.90}} \\
        \bottomrule
    \end{tabular}
    }
    \vspace{-8pt}
    \caption{\textbf{Evaluation per quality level on quality split based on size of semantics.} Best values per evaluation level are in \textbf{bold}.}
    \label{tab:result_new_allocation}
    \vspace{-10pt}
\end{table}

\vspace{1mm} \noindent \textbf{Semantic Evaluation with GT Pose and Semantics.}
Both geometric and semantic evaluations on HSSD \cite{khanna2023hssd} are shown in Table \ref{tab:abl_full}. Similar to the result of using estimated camera poses and 2D semantic segmentation, both versions of \project{} have similar performance to the corresponding fixed-size method \cite{oleynikova2017voxblox} and outperform Multi-TSDFs \cite{schmid2022panoptic} by a clear margin. 
Since we incrementally integrate semantic information by bundled raycasting, the semantic accuracy drops from 100\% even with GT 2D segmentation. A similar observation can be found in \cite{rosinol2020kimera}.

\begin{table}[ht!]
    \footnotesize
    \centering
    \resizebox{0.47\textwidth}{!}{
    \begin{tabular}{c|c|ccccc}
        \toprule
        & \textbf{Method} & \makecell{\textbf{Compl.} \\ \textbf{Error} (cm) $\downarrow$} & \makecell{\textbf{Compl. $<$5cm} \\ \textbf{Ratio} (\%) $\uparrow$} & \makecell{\textbf{Geom.} \\ \textbf{Error} (cm) $\downarrow$} & \makecell{\textbf{Sem.} \\ \textbf{Acc.} (\%) $\uparrow$} & \makecell{\textbf{Sem.} \\ \textbf{mIoU} (\%) $\uparrow$} \\  
        \midrule 
        \parbox[t]{2mm}{\multirow{4}{*}{\rotatebox[origin=c]{90}{\textit{\red{Eval @1cm}}}}} 
        &Voxblox \cite{oleynikova2017voxblox} [1cm] & \red{0.29 $\pm$ 0.22} & \red{\textbf{99.99\%}} & \red{\textbf{0.36 $\pm$ 0.37}} & \red{89.40}  & \textbf{\red{59.66}} \\
        &Multi-TSDFs~\cite{schmid2022panoptic} & \red{0.34 $\pm$ 0.56} & \red{99.71\%} & \red{0.79 $\pm$ 1.62}  & \red{78.19}  & \red{39.90} \\
        &\textbf{\project{-S}} & \red{\textbf{0.27 $\pm$ 0.22}} & \red{\textbf{99.99\%}} & \red{0.37 $\pm$ 0.42}  & \textbf{\red{89.44}}  & \textbf{\red{59.66}} \\
        &\textbf{\project{-SG}} & \red{0.29 $\pm$ 0.23} & \red{\textbf{99.99\%}} & \red{0.37 $\pm$ 0.41}  & \textbf{\red{89.44}}  & \textbf{\red{59.66}} \\
        \midrule 
        \parbox[t]{2mm}{\multirow{4}{*}{\rotatebox[origin=c]{90}{\textit{\green{Eval @4cm}}}}} 
        &Voxblox \cite{oleynikova2017voxblox} [4cm] & \green{0.92 $\pm$ 1.15} & \green{98.55\%} & \green{1.96 $\pm$ 1.96}  & \green{84.04}  & \green{61.66} \\
        &Multi-TSDFs~\cite{schmid2022panoptic} & \green{0.99 $\pm$ 2.39} & \green{96.85\%} & \green{1.57 $\pm$ 2.20} & \green{36.89}  & \green{17.80} \\
        &\textbf{\project{-S}}  & \green{0.85 $\pm$ 1.13} & \green{98.66\%} & \green{1.83 $\pm$ 1.93} & \textbf{\green{84.14}}  & \textbf{\green{61.69}} \\
        &\textbf{\project{-SG}} &\green{\textbf{0.49 $\pm$ 0.64}} & \green{\textbf{99.83\%}} & \green{\textbf{0.90 $\pm$ 1.33}} & \textbf{\green{84.14}}  & \green{61.68} \\
        \midrule 
        \parbox[t]{2mm}{\multirow{4}{*}{\rotatebox[origin=c]{90}{\textit{\blue{Eval @8cm}}}}} 
        &Voxblox \cite{oleynikova2017voxblox} [8cm] & \blue{1.43 $\pm$ 1.77} & \blue{95.54\%} & \blue{1.42 $\pm$ 2.10} & \textbf{\blue{91.14}}  & \textbf{\blue{60.89}} \\
        &Multi-TSDFs~\cite{schmid2022panoptic} & \blue{0.87 $\pm$ 1.92} & \blue{98.13\%} & \blue{0.75 $\pm$ 1.47} & \blue{78.99}  & \blue{19.31} \\
        &\textbf{\project{-S}} & \blue{1.32 $\pm$ 1.64} & \blue{96.33\%} & \blue{1.24 $\pm$ 1.90} & \blue{91.09}  & \blue{60.23} \\
        &\textbf{\project{-SG}} & \blue{\textbf{0.85 $\pm$ 1.13}} & \blue{\textbf{98.90\%}} & \blue{\textbf{0.73 $\pm$ 1.18}} & \blue{91.08}  & \blue{60.25} \\
        \bottomrule
    \end{tabular}
    }
    \vspace{-3pt}
    \caption{\textbf{Ablation study on using GT poses and semantics.}
    Best values per evaluation level are in \textbf{bold}.
    }
    \label{tab:abl_full}
    \vspace{0pt}
\end{table}

\begin{table}[ht!]
    \footnotesize
    \centering
    \resizebox{0.98\columnwidth}{!}{
    \begin{tabular}{c|ccc|c}
        \toprule
        \textbf{Method} & \makecell{\textbf{Compl.} \\ \textbf{Error} (cm) $\downarrow$} & \makecell{\textbf{Compl. $<$5cm} \\ \textbf{Ratio} (\%) $\uparrow$} & \makecell{\textbf{Geom.} \\ \textbf{Error} (cm) $\downarrow$} & \makecell{\textbf{Run Time} (ms) \\ Update TSDF $\downarrow$}\\ 
        \midrule
        \multirow{3}{*}{\textbf{Adaptive Raycasting}}
        & \textbf{\red{0.29 $\pm$ 0.23}} & \red{\textbf{99.99\%}} & \red{0.37 $\pm$ 0.41} & 
        \multirow{3}{*}{70.22 $\pm$ 11.57}
        \\
        &\green{0.49 $\pm$ 0.64} & \green{99.83\%} & \green{0.90 $\pm$ 1.33} & \\
        & \blue{\textbf{0.85 $\pm$ 1.13}} & \blue{\textbf{98.90\%}} & \blue{0.73 $\pm$ 1.18} & \\
        \midrule
        \multirow{3}{*}{\makecell{Fast Raycasting \cite{oleynikova2017voxblox} \\ {[coarse]}}}
        & \red{0.40 $\pm$ 0.31} & \red{99.98\%} & \red{0.38 $\pm$ 0.35} &
        \multirow{3}{*}{\textbf{47.05 $\pm$ \phantom{1}8.14}}
        \\
        & \green{\textbf{0.47 $\pm$ 0.54}} & \green{\textbf{99.89\%}} &
        \green{\textbf{0.84 $\pm$ 1.30}}  &\\
        & \blue{0.86 $\pm$ 1.14} & \blue{98.87\%} & \blue{\textbf{0.72 $\pm$ 1.19}} &\\
        \midrule 
        \multirow{3}{*}{\makecell{Fast Raycasting \cite{oleynikova2017voxblox} \\ {[fine]}}}
        & \textbf{\red{0.29 $\pm$ 0.23}} & \textbf{\red{99.99\%}} & \textbf{\red{0.37 $\pm$ 0.39}} &
        \multirow{3}{*}{88.03 $\pm$ 12.83}
        \\
        & \green{0.66 $\pm$ 0.82} & \green{99.53\%} & \green{0.99 $\pm$ 1.37} & \\
        & \blue{1.51 $\pm$ 1.69} & \blue{96.83\%} & \blue{1.15 $\pm$ 1.58} &\\
        \bottomrule
    \end{tabular}
    }
    \vspace{-6pt}
    \caption{\textbf{Ablation study on adaptive raycasting.} Evaluation on regions of \red{fine quality (1cm)} is in \red{red}, \green{middle quality (4cm)} is in \green{green}, and \blue{coarse quality (8cm)} is in \blue{blue}. 
    Best values per evaluation level are in \textbf{bold}.
    }
    \label{tab:abl_raycast}
    \vspace{-6pt}
\end{table}

\begin{figure}[htb]
    \centering
    \includegraphics[width=0.99\columnwidth]{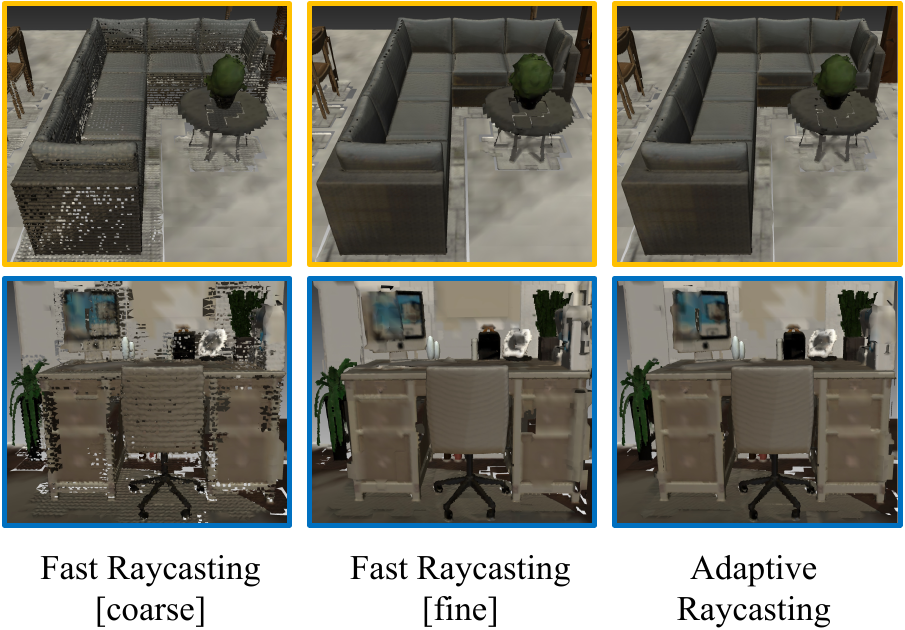}
    \caption{\textbf{Adaptive raycasting.} We show the reconstructed map of \project{-SG} with adaptive raycasting and fast raycasting \cite{oleynikova2017voxblox} in the coarse and fine resolutions.}
    \label{fig:ablation_raycasting}
    \vspace{-5pt}
\end{figure}

\vspace{1mm} \noindent \textbf{Adaptive Raycasting.}
Additional ablation study on adaptive raycasting is shown in Table \ref{tab:abl_raycast}. The experiments are conducted with GT poses and semantics on the HSSD dataset \cite{khanna2023hssd}. \textbf{Fast Raycasting [fine]} is the result of \project{-SG} when using the fast raycasting in \cite{oleynikova2017voxblox} with a single virtual grid of size 0.5 cm, which corresponds to the one used in fixed-size Voxblox (1 cm) -- by default, the virtual grid size is half the size of the voxels. \textbf{Fast Raycasting [coarse]} is the same method as \textit{w/o adaptive raycasting} in Table 4 of the main paper, which uses a single virtual grid of size 4 cm.
Compared to Fast Raycasting [fine], using adaptive raycasting (as in \project{-SG}) takes less time to update the TSDF map given each observed frame, while achieving similar performance in geometric metrics in fine quality regions. In middle and coarse quality regions, Fast Raycasting [fine] even degrades the geometric performance.
Additionally, we visualize the reconstructed map using adaptive raycasting and the two single-grid raycasting methods \cite{oleynikova2017voxblox} in Figure \ref{fig:ablation_raycasting}. With Fast Raycasting [coarse], points containing useful information for finest-resolution voxels are regarded redundant and skipped, which leads to holes in regions of fine-quality semantics. Compared to Fast Raycasting [fine], adaptive raycasting yields reconstructions with consistent quality across all regions. The additional points used in updating voxels of coarser resolution (4cm and 8cm) by Fast Raycasting [fine] are redundant and can be skipped without degenerating the reconstruction.

\begin{figure}[htb]
    \centering
    \includegraphics[width=0.9\columnwidth]{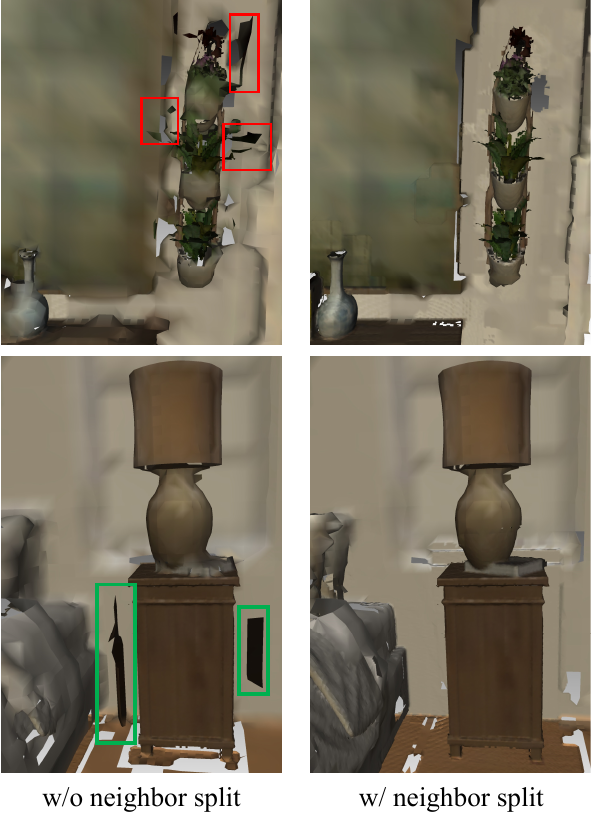}
    \caption{\textbf{Neighbor splitting.} We show the reconstructed map of \project{-SG} with and without neighbor splitting. Ghost meshes are highlighted in the color boxes.}
    \label{fig:ablation_neighbor}
    \vspace{-5pt}
\end{figure}

\vspace{1mm} \noindent \textbf{Neighbor Splitting.}
In Figure \ref{fig:ablation_neighbor}, we show the reconstructed map by \project{-SG} with and without splitting the neighbors. Without splitting the neighboring voxels, the voxels in the free space, which -- as all voxels in the map -- are initialized in the coarse resolution (see main paper), may obtain over added observations an inaccurate TSDF value, hence forming ghost meshes at the regions close to the finer-resolution voxels. The ghost meshes generated around the plant and the cabinet when no neighbors are split are highlighted in Figure \ref{fig:ablation_neighbor}.

\vspace{1mm} \noindent \textbf{Geometric Complexity.}
As mentioned in Section \ref{sec:implementation}, we project a subset of points from the 2D depth map using a stride of 2 and calculate the change of curvature \cite{rusu2010semantic,pauly2002efficient} only on these points. In this section, we study the effect of this approximation by comparing the accuracy of geometric complexity over different numbers of strides (1, 2, and 4).
In our implementation, we use the estimated geometric complexity of the coarse-resolution voxels to decide whether to split them into finer resolutions.
Therefore, whether a coarse-resolution voxel is labeled based on its geometric complexity at the correct resolution level is more crucial to the success of our system than the absolute value of geometric complexity.
To assess how accurately we estimate the geometric complexity, we evaluate the accuracy of it for different quality levels (fine, middle, coarse). 
For a quality level $r \in {fine, middle, coarse}$, the accuracy is calculated by $|S_{r_{est}} \cap S_{r_{gt}}|/|S_{r_{est}}|$, where $S_{r_{est}}$ is the set of coarse-size voxels whose estimated geometric complexity is classified as $r$ and $S_{r_{gt}}$ is the set of voxels of the coarse size whose ground truth geometric complexity is in $r$. We calculate the ground truth geometric complexity of a voxel by measuring the change of curvature from the spherical neighboring points in the aggregated projected point clouds over all frames. 

As can be observed in Table \ref{tab:geo_complexity}, our implementation (stride = 2) has similar accuracy to using all points in a frame (stride = 1), while reducing more than 10 times the processing time per frame. Although a further approximation (stride = 4) can decrease even more the processing time, the accuracy drops by about 2\% in all quality levels.

\begin{table}[ht!]
    \footnotesize
    \centering
    \begin{tabular}{c|c|ccc}
        \toprule
         \multirow{2}{*}{\textbf{Stride}} & \multirow{2}{*}{\makecell{\textbf{Process Time} \\\textbf{per Frame (ms)} $\downarrow$ }}& \multicolumn{3}{c}{\textbf{Accuracy} (\%) $\uparrow$} \\
         & & \red{Fine} & \green{Middle} & \blue{Coarse} \\
         \midrule \midrule
         1 & 1423.71 & 69.16 & \textbf{47.49} & \textbf{90.61} \\
         2 & \phantom{1}102.97 & \textbf{69.89} & 46.91 & 90.07 \\
         4 & \textbf{\phantom{1}\phantom{1}13.42} & 67.04 & 44.46 & 88.44 \\
        \bottomrule
    \end{tabular}
    \caption{\textbf{Run time and accuracy of geometric complexity estimation with different numbers of stride.} Best per level in bold. }
    \label{tab:geo_complexity}
\end{table}

\end{document}